\documentclass{article}
\usepackage{microtype}
\usepackage{graphicx}
\usepackage{subcaption}
\usepackage{booktabs} % for professional tables

\usepackage[dvipsnames]{xcolor}
\usepackage{enumitem}
\usepackage{svg}
\usepackage{bm}
\usepackage{soul}
\usepackage{setspace}

\usepackage{hyperref}

\usepackage{tikz}
\usepackage{pgfplots}
\usetikzlibrary{positioning, fit, shapes.geometric, bending, decorations.text, matrix, intersections, pgfplots.groupplots}
\usepgfplotslibrary{fillbetween}
\pgfplotsset{compat=1.18} % previously 1.15

\definecolor{C0}{HTML}{3064B8}
\definecolor{C1}{HTML}{9F6ABF}
\definecolor{C2}{HTML}{D94848}
\definecolor{C3}{HTML}{E88E3B}

\pgfplotscreateplotcyclelist{mygradient}{
{C0}, {C1}, {C2}, {C3}
}

\pgfplotsset{
try min ticks=3,
legend image code/.code={
\draw[mark repeat=2,mark phase=2]
plot coordinates {
(0cm,0cm)
(0.15cm,0cm)        %% default is (0.3cm,0cm)
(0.3cm,0cm)         %% default is (0.6cm,0cm)
};%
}
}

\usepackage[preprint]{icml2026}
\usepackage{amsmath}
\usepackage{amssymb}
\usepackage{mathtools}
\usepackage{amsthm}

\usepackage[capitalize,noabbrev]{cleveref}

\theoremstyle{plain}
\newtheorem{theorem}{Theorem}[section]
\newtheorem{proposition}[theorem]{Proposition}
\newtheorem{lemma}[theorem]{Lemma}

\theoremstyle{definition}
\newtheorem{definition}[theorem]{Definition}
\newtheorem{assumption}[theorem]{Assumption}
\theoremstyle{remark}
\newtheorem{remark}[theorem]{Remark}

\newcommand{\bct}{\bm{c}(t)}
\newcommand{\bD}{\bm{D}}
\newcommand{\bDdag}{\bD^{\dagger}}
\newcommand{\bI}{\bm{I}}
\newcommand{\bmu}{\bm{\mu}}
\newcommand{\br}{\bm{r} }
\newcommand{\brt}{\br(t)}
\newcommand{\bSigma}{\bm{\Sigma}}
\newcommand{\bs}{\bm{s}}
\newcommand{\bst}{\bs(t)}
\newcommand{\bvt}{\bm{v}(t)}
\newcommand{\but}{\bm{u}(t)}

\newcommand{\dt}{\Delta t}
\newcommand{\drt}[1]{\Delta \br_{#1}(t)}
\newcommand{\En}[1]{\mathbb{E}_{\bm{\eta}} \norm{#1}}
\newcommand{\friction}{\lambda_v}

\newcommand{\leak}{\bm{\Lambda}(t)}
\newcommand{\loss}{\mathcal{L}(t)}

\newcommand{\norm}[1]{\lVert #1 \rVert_2}
\newcommand{\score}{\nabla_{\brt} \log p(\brt)}
\newcommand{\sigmaeta}{\sigma_r \bm{\eta}(t)}
\newcommand{\sigmadt}{\sigma_r^2 \dt}
\newcommand{\timeset}[1]{ \left\{ #1 \right\}_{t=0}^T }

\DeclareMathOperator*{\argmin}{arg\,min}

\newcommand{\column}[1]{\begin{tabular}{c} #1 \end{tabular}}

\usepackage[textsize=tiny]{todonotes}

\icmltitlerunning{Leakage and Second-Order Dynamics Improve Hippocampal RNN Replay}

\begin{document}

\twocolumn[
  \icmltitle{Leakage and Second-Order Dynamics Improve Hippocampal RNN Replay}
  


  \icmlsetsymbol{equal}{*}

  \begin{icmlauthorlist}
    \icmlauthor{Josue Casco-Rodriguez}{rice}
    \icmlauthor{Nanda H. Krishna}{mila}
    \icmlauthor{Richard G. Baraniuk}{rice}
  \end{icmlauthorlist}

  \icmlaffiliation{rice}{Department of Electrical and Computer Engineering, Rice University, USA}
  \icmlaffiliation{mila}{Mila - Quebec AI Institute \& Universit\'{e} de Montr\'{e}al, Canada}

  \icmlcorrespondingauthor{Josue Casco-Rodriguez}{jc135@rice.edu}

  \icmlkeywords{RNN, replay, Langevin, sampling, hippocampus, momentum, adaptation}

  \vskip 0.3in
]



\printAffiliationsAndNotice{}  % no special notice (required even if empty)

\begin{abstract}
Biological neural networks (like the hippocampus) can internally generate ``replay'' resembling stimulus-driven activity.
Recent computational models of replay use noisy recurrent neural networks (RNNs) trained to path-integrate.
Replay in these networks has been described as Langevin sampling, but new modifiers of noisy RNN replay have surpassed this description.
We re-examine noisy RNN replay as sampling to understand or improve it in three ways:
(1) Under simple assumptions, we prove that the gradients replay activity should follow are time-varying and difficult to estimate, but readily motivate the use of hidden state leakage in RNNs for replay.
(2) We confirm that hidden state adaptation (negative feedback) encourages exploration in replay, but show that it incurs non-Markov sampling that also slows replay.
(3) We propose the first model of temporally compressed replay in noisy path-integrating RNNs through hidden state momentum, connect it to underdamped Langevin sampling, and show that, together with adaptation, it counters slowness while maintaining exploration.
We verify our findings via path-integration of 
2D triangular and T-maze paths and of high-dimensional paths of synthetic rat place cell activity.
\end{abstract}

%% ------------------------------
%% 1D Mean and Standard Deviation
%% ------------------------------
\begin{figure}[t]
\centering
\begin{tikzpicture}

\small % text size

\newcommand{\plotOU}[3]{
    \addplot[#1, #2, thick] table[x=time, y=mean_#3] {means/1d_ou.txt};
    \addplot[name path=upper, draw=none] table[x=time, y expr = \thisrow{mean_#3} + \thisrow{std_#3}] {means/1d_ou.txt};
    \addplot[name path=lower, draw=none] table[x=time, y expr = \thisrow{mean_#3} - \thisrow{std_#3}] {means/1d_ou.txt};
    \addplot[#1, nearly transparent] fill between[of=upper and lower];
}

\begin{groupplot}[
    group style = {
        group size = 4 by 1,
        horizontal sep = 0.06\linewidth},
    width = 0.38\linewidth,
    height = 0.35\linewidth,
    title style={yshift=-2mm, font=\scriptsize},
    xlabel=Time,
    xlabel shift={-1mm},
    xmin=0, xmax=2,
    ymin=-0.6, ymax=5.6,
    xtick={0, 1, 2},
    axis x line*=bottom,
    axis y line*=left,
    table/col sep=comma
    ]

\nextgroupplot[title=$s(t)$]
    \plotOU{red}{dashed}{s};

\nextgroupplot[title=$r(t)$, yticklabels={}]
    \plotOU{red}{dashed}{s};
    \plotOU{blue}{}{overdamped};

\nextgroupplot[title= $r(t)$ underdamped, yticklabels={}]
    \plotOU{red}{dashed}{s};
    \plotOU{blue}{}{underdamped};

\nextgroupplot[title= $r(t)$ w/ adaptation, yticklabels={}]
    \plotOU{red}{dashed}{s};
    \plotOU{blue}{}{adaptation};

\end{groupplot}

\node at (3.8, 2) {Mean and Std of $s(t)$ (awake paths) and $r(t)$ (replay paths)};

\end{tikzpicture}
\caption{
\textbf{Underdamped dynamics accelerate offline replay, adaptation slows it.}
Here we simulate a noisy RNN $r(t)$ that optimally path-integrates an Ornstein-Uhlenbeck process $s(t)$ from its velocity $s'(t)$.
We assume $r(t)$ minimizes the loss in \Cref{eq:loss_upper} and thus evolves according to its \textit{score function} $\nabla_{r(t)} \log p(r(t))$ (Equations \ref{eq:rnn_langevin} and \ref{eq:score_ou_main}), performing a variant of Langevin sampling when no input is given.
Above, we compare three modifiers of RNN activity: the default (no modification, a.k.a. \textit{overdamped}), our proposed \textit{underdamped} (momentum), and \textit{adaptation} (negative feedback) dynamics.
Each modifier affects the replay distribution $p(r(t))$ in different ways:
underdamped sampling accelerates $p(r(t))$ towards $p(s(t))$,
decreasing the distance between them, while adaptation slows convergence of $p(r(t))$ towards $p(s(t))$, increasing this distance.
}
\label{fig:1d_mean_std}
\end{figure}
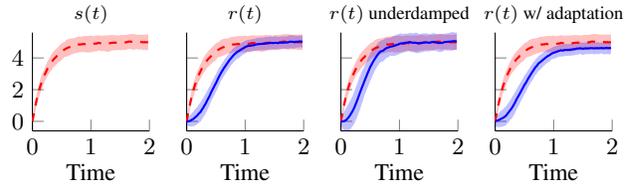
\vspace{-0.5em}
\section{Introduction}

During quiescent periods such as sleep or wakeful resting, some neural circuits internally generate activity resembling that of active periods \citep{tingley2020methods}. Such ``replay'' phenomena have been observed in the prefontal \citep{euston2007fast, peyrache2009replay}, sensory \citep{kenet2003spontaneously, xu2012activity}, motor \citep{hoffman2002coordinated}, and entorhinal cortices \citep{gardner2022toroidal}; the anterior thalamus \citep{peyrache2015internally}; and the hippocampus \citep{hippocampal_sharp_waves, replay_of_neuronal, nadasdy1999replay, lee2002memory, foster2017replay}.
Of these circuits, the hippocampus is particularly interesting because its robustness in tasks like navigation \citep{hippocampus_cognitive_map, model_hippocampal, deciphering_hippocampal} and planning \citep{hippocampal_place-cell_sequences, dorsal_hippocampus} during active states seems crucially tied to its spontaneous activity during quiescent states \citep{two-stage_model, hippocampal_sharp_wave_ripple, sleep_price, olafsdottir2015hippocampal,olafsdttir2018therole}.

While some works have produced replay using supervised generative models \citep{deperrois2022learning}, most existing models of hippocampal activity treat replay as an emergent byproduct of careful network design. Relevant network parameters include connectivity structures \citep{shen1996modeling, milstein2023offline}, local plasticity mechanisms \citep{hopfield2009neurodynamics, litwin2014formation, theodoni2018theta, haga2018recurrent, predictive_learning_rules}, firing rate adaptation \citep{chu2023firing, computational_model_preplay, cell_assembly_sequences, noisy_adaptation, dynamics_of_adaptive_canns}, and input modulation \citep{kang2019replay}. While these models reproduce aspects of replay, they are typically motivated by empirical findings and lack rigorous theoretical justification.

A more principled model of hippocampal function with emergent replay is sequential predictive learning \citep{sufficient_conditions, sequential_predictive_learning}, wherein neural circuits predict dynamic environment or task variables from imperfect observations thereof, e.g., path-integrating velocity measurements to track a position.
This normative description of the hippocampus as a sequence predictor \citep{computational_approach,hippocampus_predictive_map} matches hippocampal encodings of upcoming stimuli \citep{hippocampus_preserves_order} and prediction errors \citep{hippocampal_representations_switch, emergence_predictive_model}, and neural activity sweeps that represent possible future trajectories \citep{constant_sub-second, neural_ensembles_ca3}.
Unlike traditional, hand-crafted models of hippocampal circuits (i.e., continuous attractor networks \citep{path_integration, attractor_neural_networks_storing}), sequential predictive learning models are trained from data.
Nonetheless, they account for the emergence of place cells \citep{predictive_learning_network_mechanism, sequential_predictive_learning, predictive_sequence_learning}, grid cells \citep{cueva2018emergence, sorscher2019unified}, and head direction cells \citep{Cueva2020Emergence, Uria2020Spatial}; can incorporate phenomena like theta oscillations \citep{sequential_predictive_learning}; and exhibit quiescent replay activity \citep{sufficient_conditions, sequential_predictive_learning, predictive_sequence_learning}.

\citet{sufficient_conditions} provided the first theoretical foundation for replay in sequential predictive learning networks, showing analytically that they generate diffusive replay \citep{hippocampal_reactivation_random} during quiescent activity (i.e., noise-driven activity in the absence of inputs) by Langevin sampling \citep{besag1994} from the waking activity distribution using its score function. Subsequent empirical work \citep{sequential_predictive_learning} introduced new mechanisms to induce \textit{exploration} in replay through negative feedback, i.e., neural adaptation. Exploration is the notion that replay expresses a variety of behavioral sequences \citep{hippocampal_replay_extended,content_of_hippocampal}, and is associated with long trajectories in neural space, visitation of multiple attractor basins, and transitions that were not present in awake activity. Adaptation can destabilize attractors and induce sudden transitions in replay activity \citep{cell_assembly_sequences,noisy_adaptation,dynamics_of_adaptive_canns,sequential_predictive_learning}, thereby facilitating exploration, and is thought to play a key role in the dynamics of replay \textit{in vivo} \citep{nrem_sleep}. However, existing theory on sequential predictive learning cannot account for these mechanisms. Furthermore, sequential predictive learning models do not currently account for the \textit{temporal compression} of replay sequences relative to awake sequences of activity \citep{nadasdy1999replay, hippocampal_sharp_wave_ripple, speed_time_compressed, emergence_preconfigured}.
This phenomenon, which could be caused by short-term facilitation \citep{temporal_compression_mediated,phase_precession_neural_code}, is not currently captured in replay from any trained RNN model to our knowledge.

Overall, while sequential predictive learning is a promising model of hippocampal function and replay, its recent empirical advances have outpaced its theoretical foundations. Moreover, it is unclear how to incorporate phenomena like temporal compression in these models, or how inductive biases in RNN design affect replay. We remedy these shortcomings by characterizing how RNN design and Langevin sampling statistics affect each other.
Some results describe how RNN design affects the speed and variance of replay activity, while others start from the Langevin sampling formulation of replay and either explain existing architectural choices as useful inductive biases, or propose new mechanisms to again modulate sampling.
In summary, we answer three questions:
\begin{enumerate}
\item Optimal path-integration in the presence of noise requires RNNs to learn the score function of the noisy activity distribution.
What is this function, might it inform RNN design?
\textit{The score function is time-variant and difficult to estimate, even for simple distributions, but our expression of it motivates the addition of leakage (linear dynamics) in RNNs.}
\item Adaptation (negative feedback) empirically induces exploration in replay.
How does it affect Langevin sampling?
\textit{Adaptation induces non-Markov second-order Langevin sampling that destabilizes attractors, which can both help (diversify) and hurt (slow) replay.}
\item Traditional generative models benefit from a wide array of sampling techniques. Could new sampling methods improve replay in noisy RNNs?
\textit{Underdamped Langevin sampling via momentum quickens neural replay, like temporal compression induced by short-term facilitation} in vivo\textit{, and mitigates slowing from adaptation while maintaining exploration.}
\end{enumerate}

\section{Background}

In this section, we provide background and summarize results from prior work \citep{sufficient_conditions} that has described replay in path-integrating neural circuits as the sampling during quiescence (i.e., the absence of any inputs) of neural activity states from the distribution of waking, task-like activity. In short, the dynamics of recurrent networks that learn to optimally path-integrate noisy inputs cause the network's states even in the absence of inputs to resemble those attained during actual task performance. We also provide details on mechanisms used in the literature to improve or bias replay, which we explore the effects of in more detail in this work. For an overview of this section, see \Cref{appendix:background_overview}.

\subsection{Langevin Dynamics}

Langevin sampling from an unknown distribution $p(x)$ entails stochastic gradient ascent of an iterate $x(t)$ along the log-likelihood of $p(x)$ via its \textit{score function} $\nabla \log p(x)$ or an estimate thereof:
\begin{equation} \label{eq:overdamped_langevin}
    \dot{x}(t) = x(t) + \nabla \log p(x) + \sqrt{2} \eta(t),
\end{equation}
where $\eta(t)$ is Gaussian white noise. While Equation \ref{eq:overdamped_langevin} describes \textit{overdamped} dynamics, there also exist \textit{underdamped} Langevin dynamics\footnote{Discretized underdamped Langevin dynamics are a form of Hamiltonian MCMC \citep{underdamped_langevin_mcmc}.} (Equations \ref{eq:underdamped_langevin_2nd_order} or \ref{eq:underdamped_langevin_1st_order}, see Chapter 6 of \citet{stochastic_processes}) that converge faster to the target distribution $p(x)$ and better utilize noisy gradients \citep{underdamped_langevin_mcmc}:
\begin{gather} 
    \ddot{x}(t) = \nabla \log p(x) - \gamma \dot{x}(t) + \sqrt{2 \gamma} \eta(t), \quad \mathrm{or}
    \label{eq:underdamped_langevin_2nd_order}
    \\
    \dot{x}(t) = v(t), \enspace \dot{v}(t) = \nabla \log p(x) - \gamma v(t) + \sqrt{2 \gamma} \eta(t)
    \label{eq:underdamped_langevin_1st_order}
\end{gather}

In this work, we consider noisy RNNs that have implicitly learned to perform Langevin sampling of their own, fixed distributions of activity during task performance, when driven by just intrinsic noise and in the absence of inputs. That is, the activity of the RNNs at each unrolled timestep in the absence of inputs represents a plausible and likely vector of network activity during actual task performance in the presence of inputs. In particular, we view such networks in the context of replay, where neural circuits recapitulate task-like activity even during sleep.

\subsection{Offline Replay in RNNs}

This work focuses on RNNs that implicitly learn to act as generative models over a fixed distribution of input sequences.
\citet{sufficient_conditions} have shown how noisy RNNs trained to \emph{path-integrate} their inputs implicitly learn statistics that produce Langevin sampling of their own task-relevant activity distribution when no input is given, demonstrating statistically faithful replay sequences during quiescence. That is, the RNNs' activity in the absence of inputs ``replays'' states from the same distribution as RNN activity with inputs during actual task performance. This leads to the generation or ``replay'' of output sequences resembling those that the network sees during training or task performance with inputs.
Path-integration is particularly relevant to neuroscience: animals can leverage motion cues, observations, or prior experiences to accurately estimate positions \citep{neural_dynamics_landmark, human_path_integration}, and neural circuits like the entorhinal cortex \citep{sorscher2019unified} have been identified to perform such computations.
Here we summarize the finding that noisy RNNs trained to path-integrate input time-series learn the \emph{score function} of the input distribution \citep{sufficient_conditions}.

\begin{definition}
A \emph{noisy recurrent neural network (RNN)} has hidden states $\brt$ that evolve at each timestep $t$ via some (nonlinear) function of its previous hidden states, an input signal $\but$, and noise:
\begin{equation}
    \label{eq:rnn_definition}
    \br(t + \dt) = f(\brt, \but, \sigmaeta)
\end{equation}
\end{definition}

\begin{definition}
A \textit{path-integration} objective $\loss{}$ penalizes the difference between a state variable $\bst$ and a learnable linear projection of the RNN hidden state $\brt$:
% at every timestep $t$:
\begin{equation}
    \loss = \En{\bst - \bD\brt}
\end{equation}
\end{definition}

\begin{assumption}
\label{assumption:additive_decomposition}
\citet{sufficient_conditions} approximate the hidden state dynamics of a noisy RNN as the sum of two functions\footnote{
Each function $\drt{1}, \drt{2}$ can depend on
variables beyond $t$, but they are omitted for concision.
} and white noise $\sigmaeta \sim \mathcal{N}(0, \sigma_r^2 \dt)$.
\begin{equation}
\begin{split}
\label{eq:rnn_update_definition}
    \drt{} &= \br(t + \dt) - \brt \\
    &\approx \drt{1} + \drt{2} + \sigmaeta
\end{split}
\end{equation}
\end{assumption}

\begin{assumption}
\label{assumption:gaussian_r(t)}
The optimal $\brt$ minimizes $\loss$ such that $p(\br^*(t))$ is normal around $\bm{D}^\dagger \bst$:
\begin{equation}
    p(\br^*(t) | \bst) \sim \mathcal{N}(\bDdag \bst, \, \bI \sigmadt)
    \label{eq:p_r|s}
\end{equation}
\end{assumption}

\begin{lemma}
\label{lemma:upper_bound}
With Assumptions \ref{assumption:additive_decomposition} and \ref{assumption:gaussian_r(t)},  $\mathcal{L}(t+\Delta t)$ is upper bounded by $\widehat{\mathcal{L}}(t+\Delta t)$:
\begin{equation}
\begin{split}
    \label{eq:loss_upper}
    \widehat{\mathcal{L}}(t+\Delta t) &= \lVert \bs'(t) - \bD \drt{2} \rVert + \En{\bm{D}\sigma_r\boldsymbol{\eta}(t)}
    \\
    &+ \lVert \bm{D} \rVert_F \En{ \drt{1} + \sigma_r\boldsymbol{\eta}(t-\Delta t)}
\end{split}
\end{equation}
\end{lemma}

\begin{assumption}
\label{assumption:greedy}
The optimal $\brt$ greedily minimizes $\widehat{\mathcal{L}}$ at $t$ only, without concern for long-range dependencies:
\begin{equation}
\begin{split}
    \timeset{\br^*(t)} &= \argmin_{\timeset{\brt}} \int_{t=0}^T \mathcal{L}(t+\Delta t) \Delta t
    \\
    &\approx \biggl\{ \argmin_{\brt} \widehat{\mathcal{L}}(t+\Delta t) \biggr\}_{t=0}^T
\end{split}
\end{equation}
\end{assumption}

\begin{theorem}
\label{theorem:optimal_update}
Given Assumption \ref{assumption:greedy}, the optimal update 
$\Delta \br^*(t) = \argmin_{\drt{2}} \widehat{\mathcal{L}}(t+\Delta t) + \argmin_{\drt{1}} \widehat{\mathcal{L}}(t+\Delta t) + \sigmaeta$
follows $\bs'(t)$ and the score function of $p(\brt)$:
\begin{equation}
    \Delta \br^*(t) =\bD^\dagger \bs'(t) \dt + \sigma_r^2 \dt \score + \sigmaeta
    \label{eq:delta_r_1_and_2_star}
\end{equation}
\end{theorem}

During training, noisy RNNs are (unless otherwise stated, see \Cref{sec:bias_methods}) provided $\bm{u}(t) = \bs'(t)$ (or a nonlinear observation thereof) to perform path-integration and focus on denoising.
\begin{theorem}
\label{theorem:langevin_replay}
In the absence of input (quiescence), a noisy RNN already trained to path-integrate $\bst$ from $\bs'(t)$ will perform gradient ascent along the score function of $\brt$.
If $p(\brt)$ and $p(\bst)$ are stationary, then this ascent is Langevin sampling (Equation \ref{eq:overdamped_langevin}).
If the variance of $\sigmaeta$ is scaled by a factor of 2, then $p(\brt)$ is guaranteed to converge to the steady-state distribution $p(\br) = p(\bD^\dagger \bs)$.
\begin{equation}
\begin{split}
    \label{eq:rnn_langevin}
    &\text{If } \drt{} = \sigma_r^2 \dt \score + \sqrt{2} \sigmaeta, \\
    &\text{then} \lim_{t \rightarrow \infty} p(\br(t)) = p(\bD^\dagger \bs)
\end{split}
\end{equation}
\end{theorem}

\subsection{Existing Methods of Biasing Replay in RNNs}
\label{sec:bias_methods}

\paragraph{Neural adaptation.}
Biological neurons can mitigate prolonged or low-frequency activity via negative feedback, or \textit{adaptation} \citep{neural_adaptation, spike_frequency_adaptation}.
This feedback has proven important for describing \textit{in vivo} hippocampal activity and replay \citep{cell_assembly_sequences, nrem_sleep}, and in computational models of replay has been shown to encourage long replay trajectories (exploration) by preventing neural activations from getting stuck in attractor basins \citep{noisy_adaptation, dynamics_of_adaptive_canns, sequential_predictive_learning}.
Like \citet{sequential_predictive_learning}, we define adaptation as negative feedback $\bct$ added to RNN activity $\brt$ (Equation \ref{eq:rnn_definition}) after training\footnote{Subtraction of the moving average $\bct$ also arises naturally from greedy minimization of $\loss + \frac{1}{2} \norm{\bct}^2$.}:  \begin{equation}
\begin{split}
\label{eq:adaptation_definition}
\br(t + \dt) &= f(\brt, \but, \sigmaeta) - \bct, \\ 
\Delta \bct &= \frac{1}{\tau_a} (-\bct + b_a \brt)
\end{split}
\end{equation}

\paragraph{Masked training.}
Denoisers and autoencoders benefit from \textit{masked training}, wherein some regions of input data are set to zero before model processing \citep{survey_on_masked}. %, masked_feature}.
\citet{sequential_predictive_learning} introduce masked training for path-integration by periodically masking the input $\but$ (the observation of $\bs'(t)$): only at every $k$-th timestep does the RNN observe a nonzero input (Equation \ref{eq:masked_training}).
\begin{equation}
    \label{eq:masked_training}
    \but = \begin{cases}
        \bs'(t), & t \text{ mod } k = 0 \\
        \bm{0}, & \mathrm{otherwise}
    \end{cases}
\end{equation}
\citet{sequential_predictive_learning} found that masked training makes replay sequences more coherent and makes manifolds of neural activity more similar to the spatial layout of the environment.
We found that masked training improves replay stability, so we use it (with $k \geq 3$) in training all RNNs.
%\section{Demystifying the RNN Score Function \josue{new title?}}
\section{Estimating the Score Function of Noisy RNN Activity}
\label{sec:rnn_score}

Noisy RNNs trained to path-integrate implicitly learn the score function of their activity.
Previous works have not examined the score function; in fact, they assume the distribution of RNN activity $p(\brt)$ is stationary (\Cref{theorem:langevin_replay}), and thus the score function $\score$ depends only on $\brt$ \citep{sufficient_conditions}.
However, we refute this assumption, and in doing so reveal the role of leakage
in path-integration:
%in Langevin sampling: 
even if the RNN path-integrates a simple Gaussian process, the score function requires information beyond $\brt$, which it employs through linear leakage (decay) of $\brt$.
This linearity suggests that leakage is useful for path-integration, which we confirm experimentally.

\subsection{Challenges in Simple Distributions}
\label{sec:challenges}

First, we examine how the score function of optimal path-integrating RNN activity $\brt$ has nonstationarities that are challenging to perfectly estimate, even for simple Gaussian processes.

\begin{assumption}
\label{assumption:gaussian_process}
The observed states $\bst$ form some Gaussian process: $p(\bst) \sim \mathcal{N} \left( \bm{\mu}_{\bst}, \bm{\Sigma}_{\bst} \right)$.
\end{assumption}
\begin{theorem}
\label{theorem:gaussian_score_linear}
With Assumption \ref{assumption:gaussian_process}, the score function of trained activity $\brt$ has a closed form which, while linear in $\brt$, is nonlinear with respect to the parameters of $p(\bst)$ (see \Cref{sec:appendix_score_of_r(t)}):
\begin{equation}
\label{eq:general_gaussian_score}
\sigmadt \, \score = - \leak \left( \brt - \bDdag \bmu_{\bst} \right)
\end{equation}
\end{theorem}
With $\leak = \sigmadt \left( \bI \sigmadt + \bDdag \bSigma_{\bst} (\bDdag)^T \right)^{-1}$ as the \textit{leakage matrix} of $\brt$, we can already gain some insight from Equation \ref{eq:general_gaussian_score}.
\begin{remark}
\label{remark:score_linearity}
If $p(\bst)$ is Gaussian, then the score of $p(\brt)$ is simply a linear function of $\brt$, but its parameters are nonlinear functions of time, and only as stationary as $p(\bst)$.
\end{remark}
\begin{remark}
\label{remark:leakage_eigs}
The eigenvalues of the leakage matrix $\leak$ are always between 0 and 1 (see Appendix \ref{sec:appendix_score_of_r(t)}).
Moreover, the eigenvalues of $\leak$ and $\bDdag \bSigma_{\bst} (\bDdag)^T$ inversely correlate: strong decay of $\brt$ implies weak noise $\bSigma_{\brt}$, and weak decay of $\brt$ implies strong noise $\bSigma_{\brt}$.    
\end{remark}

To further illustrate how estimating the score function of $\brt$ (Equation \ref{eq:general_gaussian_score}) can be challenging, we now examine a scalar RNN $r_{ou}(t)$ trained to path-integrate Ornstein-Uhlenbeck processes.
\begin{assumption}
\label{assumption:ou}
The observed states follow a scalar Ornstein-Uhlenbeck process parameterized by $\theta, \mu, \sigma_s$:
$s_{ou}'(t) = \theta (\mu - s_{ou}(t)) + \sigma_s \eta(t)$, where $p(s(0)) \sim \mathcal{N}(0, \sigma_0^2)$. 
In other words, $p(s_{ou}(t))$ parameterizes a directed random walk from $s_{ou}(0)$ to $\mu$.
\end{assumption}
\begin{remark}
\label{remark:score_ou}
The score function of the optimal $r_{ou}(t)$ under Assumption \ref{assumption:ou} follows from \Cref{eq:gaussian_score} (see \Cref{appendix:more_score}):
\begin{multline}
\label{eq:score_ou_main}
\sigmadt \, \nabla_{r_{ou}(t)} \log p(r_{ou}(t)) = \\
\frac{
-\sigmadt \left( r_{ou}(t) - \mu \left(1 - e^{-\theta t} \right) \right)
}{
\sigmadt + \frac{\sigma_s^2}{2 \theta} \left( 1 - e^{-2 \theta t} \right) + \sigma_0^2 e^{-\theta t}
}
\end{multline}
\end{remark}

The score function, and thus the optimal quiescent activity, is evidently complex and nonstationary: $\lim_{t \rightarrow 0} \sigmadt \nabla_{r_{ou}(t)} \log p(r_{ou}(t)) = -\frac{\sigmadt}{\sigmadt + \sigma_0^2} r_{ou}(t)$, while $\lim_{t \rightarrow \infty} \sigmadt \nabla_{r_{ou}(t)} \log p(r_{ou}(t)) = -\frac{\sigmadt}{\sigmadt + \sigma_s^2/2 \theta}(r_{ou}(t) - \mu)$.
While one could force stationarity by implicitly assuming the process starts at $s_{ou}(0)=\mu$, or assuming the steady-state dynamics ($t \rightarrow \infty$) are the most important, we argue that any such approach would miss a fundamentally relevant aspect of the Ornstein-Uhlenbeck process for navigation: \textit{intention}.
Unlike the Wiener process (\Cref{appendix:more_score}), the Ornstein-Uhlenbeck process can describe a random walk that \textit{intentionally} navigates from $s_{u}(0)$ to $\mu$, rather than one that simply wanders around $\mu$.
Thus, for navigation, the non-stationary, or ``early'', dynamics of the Ornstein-Uhlenbeck process are the most salient.
Given the relevance of non-stationary dynamics for navigation, our analyses focus on the entire course of replay dynamics (rather than steady-state distributions), examining properties like speed and path diversity (exploration).

\subsection{The Advantage of Leakage}

%% ------------------
%% Leakage loss plots
%%-------------------
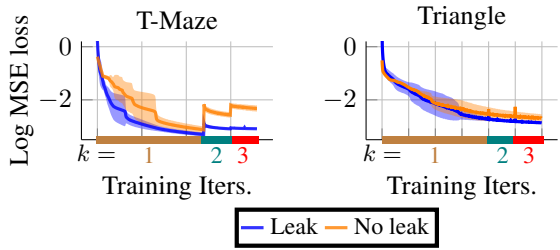
\begin{figure}[h]
\centering
\begin{tikzpicture}

\pgfplotsset{/pgfplots/group/every plot/.append style = {
    ultra thick,%, no markers
    legend style={
        legend columns=2,
        anchor=south west}
}};

\begin{groupplot}[
    group style = {
        group size = 2 by 1,
        horizontal sep = 0.15\linewidth},
    width = 0.5\linewidth,
    height = 0.35\linewidth,
    title style={yshift=-2mm,},
    axis x line*=bottom,
    axis y line*=left,
    ymax=0.2,
    ymin=-3.5,
    ylabel shift={-2mm},
    xlabel={Training Iters.},
    xlabel shift={1mm},
    xtick=\empty,
    xticklabel=\empty,
    xtick={0, 100, 200, 300, 400, 500, 600},
    grid,
    no marks,
    table/col sep=comma
    ]

    %% ------------
    %% Group plot 1
    %% ------------
    \nextgroupplot[title={T-Maze},
        ylabel={Log MSE loss},
        legend style={
            at={(0.8,-1.1)},
            inner sep=1pt,
            font=\small}
    ]
            
    \addplot[color=blue, very thick, draw opacity=0.8] table[x index=0, y={log_mean}] 
    {csv/leak_losses_tmaze.csv}; 
    \addlegendentryexpanded{Leak};    
    \addplot[color=orange, very thick, draw opacity=0.8] table[x index=0, y={noleak_log_mean}] 
    {csv/leak_losses_tmaze.csv};  
    \addlegendentryexpanded{No leak};   

    %% -------------------------------

    \addplot[name path=upper1, draw=none] table[x index=0, y expr=\thisrow{log_mean} + \thisrow{log_std}] {csv/leak_losses_tmaze.csv};

    \addplot[name path=lower1, draw=none] table[x index=0, y expr=\thisrow{log_mean} - \thisrow{log_std}] {csv/leak_losses_tmaze.csv};

    \addplot[fill=blue, fill opacity=0.4] fill between[of=upper1 and lower1];

    \addplot[name path=upper2, draw=none] table[x index=0, y expr=\thisrow{noleak_log_mean} + \thisrow{noleak_log_std}] {csv/leak_losses_tmaze.csv};

    \addplot[name path=lower2, draw=none] table[x index=0, y expr=\thisrow{noleak_log_mean} - \thisrow{noleak_log_std}] {csv/leak_losses_tmaze.csv};

    \addplot[fill=orange, fill opacity=0.4] fill between[of=upper2 and lower2];

    %% ------------
    %% Group plot 2
    %% ------------
    \nextgroupplot[title={Triangle}]
        
    \addplot[color=blue, very thick] table[x index=0, y={log_mean}] 
    {csv/leak_losses_triangle.csv};   
    \addplot[color=orange, very thick] table[x index=0, y={noleak_log_mean}] 
    {csv/leak_losses_triangle.csv};  
    
    %% -------------------------------

    \addplot[name path=upper1, draw=none] table[x index=0, y expr=\thisrow{log_mean} + \thisrow{log_std}] {csv/leak_losses_triangle.csv};

    \addplot[name path=lower1, draw=none] table[x index=0, y expr=\thisrow{log_mean} - \thisrow{log_std}] {csv/leak_losses_triangle.csv};

    \addplot[fill=blue, fill opacity=0.4] fill between[of=upper1 and lower1];

    \addplot[name path=upper2, draw=none] table[x index=0, y expr=\thisrow{noleak_log_mean} + \thisrow{noleak_log_std}] {csv/leak_losses_triangle.csv};

    \addplot[name path=lower2, draw=none] table[x index=0, y expr=\thisrow{noleak_log_mean} - \thisrow{noleak_log_std}] {csv/leak_losses_triangle.csv};

    \addplot[fill=orange, fill opacity=0.4] fill between[of=upper2 and lower2];

\end{groupplot} 

%% --------------------
%% Add unmasking colors
%% --------------------
% Plot 1
\fill[brown, name path=fo] (0.2,-0.05) rectangle (1.6,0.05);
\fill[teal] (1.6,-0.05) rectangle (2,0.05);
\fill[red] (2,-0.05) rectangle (2.35,0.05);

\node[brown] at (0.9,-0.2) {\footnotesize 1};
\node[teal] at (1.8,-0.2) {\footnotesize 2};
\node[red] at (2.15,-0.2) {\footnotesize 3};

% Plot 2
\fill[brown] (4,-0.05) rectangle (5.4,0.05);
\fill[teal] (5.4,-0.05) rectangle (5.75,0.05);
\fill[red] (5.75,-0.05) rectangle (6.15,0.05);

\node[brown] at (4.7,-0.2) {\footnotesize 1};
\node[teal] at (5.6,-0.2) {\footnotesize 2};
\node[red] at (5.95,-0.2) {\footnotesize 3};

%\node at (-0.4,-0.15) {\scriptsize Masking $k=$};
\node at (0.2,-0.17) {\footnotesize $k=$};
\node at (4,-0.17) {\footnotesize $k=$};

\end{tikzpicture}
\caption{
\textbf{Leakage helps path-integration.}
Here we train RNNs on two tasks, ablating the leakage term.
Leakage helps training, especially when losses increase with the masking difficulty $k$ (note that $k=1$ is equivalent to \emph{unmasked} training, see Equation \ref{eq:masked_training}).
Means are solid, standard deviations are faint.
}
\label{fig:leak_losses}
\end{figure}

Theorem \ref{theorem:gaussian_score_linear} and Remark \ref{remark:score_linearity} suggest that linear leakage may be a useful inductive bias for RNNs learning to path-integrate: the score function for a Gaussian process is linear with respect to $\brt$, although the parameters of said linearity are nonlinear functions of time.
We examine the utility of leakage by comparing two RNNs trained to path-integrate:
$\br(t + \dt) = \kappa \brt + f_1(\brt, \but, \sigmaeta)$ (RNN with leakage $0 < \kappa < 1$) and 
$\br(t + \dt) = f_2(\brt, \but, \sigmaeta)$ (RNN without leakage), where $f_1, f_2$ are shallow ReLU layers trained separately (see \Cref{appendix:discretization}).
The second is more reminiscent of traditional RNNs in machine learning (e.g., ReLU or gated RNNs), which do not typically employ leakage.
In \Cref{fig:leak_losses}, we show that leakage is useful for path-integration, especially with masked training ($k > 1$).
Our results suggest that leakage is a useful component of path-integrating RNNs, although other architectures may be able to successfully learn without it (e.g., the layer-norm RNN of \citet{sequential_predictive_learning}).
\section{Second-Order Langevin Sampling for Neural Replay}

In the previous section, we examined the score function of trained path-integrating RNN activity.
Now, we are assuming that trained RNNs have well-estimated the score function, and follow it during quiescent (internally-driven) activity to generate replay via Langevin dynamics.
Here we ask how modulating the dynamics of such RNNs affects the distribution of replay---in other words, how changing the RNN dynamics biases the distribution of replay.
We see that adaptation (negative feedback) incurs a non-ideal form of second-order Langevin sampling, so we propose a complementary alternative that explicitly performs underdamped second-order sampling.

\subsection{Adaptation as Underdamped Langevin Dynamics}

First we examine adaptation (negative feedback), as defined in \Cref{sec:bias_methods}.

\begin{proposition}
Adding adaptation (\Cref{eq:adaptation_definition}) to a trained path-integrating RNN during quiescence (\Cref{eq:rnn_langevin}) incurs Langevin sampling with negative feedback:
%\begin{equation}
%\begin{split}
\begin{multline}
\label{eq:rnn_langevin_adaptation}
\drt{} = \sigmadt \, \score + \sqrt{2} \sigmaeta - \bct, \\
\Delta \bct = \frac{1}{\tau_a} (-\bct + b_a \brt)
%\end{split}
%\end{equation}
\end{multline}
\end{proposition}
For the clearest illustration of the effects of adaptation, let us examine a stationary $p(\brt)$---a simplification which we argued in \Cref{sec:challenges} is not realistic, but is nonetheless intuitive.
\begin{assumption}
\label{assumption:gaussian_stationary}
The observed states are normal (Assumption \ref{assumption:gaussian_process}) and stationary: $p(\bst) \sim \mathcal{N} \left( \bm{\mu}, \bm{\Sigma} \right)$.
\end{assumption}
\begin{theorem}
Adding adaptation to an RNN trained to path-integrate states drawn from a stationary Gaussian distribution
(Assumption \ref{assumption:gaussian_stationary} and \Cref{eq:rnn_langevin_adaptation}) produces the following second-order stochastic dynamics during quiescence (see \Cref{sec:appendix_adaptation_sde}):
\begin{equation} \label{eq:r_adaptation_2nd_order}
\begin{split}
    \br''(t)
    =
    \left(
    \frac{b_a}{\tau_a} \bI + \sigmadt \frac{d^2}{d\brt^2} \log p(\brt)
    \right) \br'(t)
    \\
    -
    \frac{b_a}{\tau_a} \sigmadt~ \score
    +
    \frac{1}{\tau_a} \br(t)
    \\
    -
    \sigma \frac{b_a}{\tau_a} \bm{\eta}(t) + \sigma \bm{\eta}'(t)
\end{split}
\end{equation}
\end{theorem}

Comparing Equation \ref{eq:r_adaptation_2nd_order} with Equation \ref{eq:underdamped_langevin_2nd_order}, the two indeed resemble each other: adaptation seems to induce a form of underdamped Langevin dynamics.
This may help explain the observed utility of adaptation for generating replay \citep{cell_assembly_sequences, nrem_sleep, sequential_predictive_learning}.
Moreover, since we established in Section \ref{sec:rnn_score} that the score function of even a basic stochastic process is difficult to estimate, the effectiveness of underdamped Langevin sampling for working with noisy gradients \citep{underdamped_langevin_mcmc} may be useful when $\score$ is poorly estimated.

However, interpreting \Cref{eq:r_adaptation_2nd_order} as underdamped Langevin sampling reveals some shortcomings thereof as a sampling method.
\begin{remark}
The coefficient of $\br'(t)$ is usually constant, and should be positive to ensure convergence (\citet{stochastic_processes}, pg. 183), but $\frac{b_a}{\tau_a} \bI + \score$ is not constant and could be negative.
\end{remark}
\begin{remark}
\label{remark:adaptation_negative_sign}
Underdamped Langevin sampling from $\brt$ should not have a negative sign in front of $\score$ if the intention is to maximize $p(\brt)$. 
\end{remark}

\subsection{Replay via Explicit Underdamped Langevin Dynamics}

While adaptation is a biologically plausible way to perform a variant of underdamped Langevin dynamics in RNNs, we propose an alternative method, from Equation \ref{eq:underdamped_langevin_1st_order}, to more clearly and directly perform underdamped Langevin sampling.
It is conceptually similar to RNNs with momentum \citep{momentumrnn}: a velocity term $\bvt$ accumulates previous $\brt$ values when friction $\friction \in [0, 1]$ is below 1\footnote{
For a comparison of $\friction$ and the $\gamma$ term from Equations \ref{eq:underdamped_langevin_1st_order} and \ref{eq:underdamped_langevin_2nd_order}, see \Cref{appendix:deviations_langevin}.
}.
\begin{definition}
We implement \textit{explicitly underdamped} dynamics via momentum governed by friction $\friction$.
When $\friction = 1$, dynamics revert to overdamped (\Cref{eq:rnn_definition}):
\begin{multline}
    \label{eq:underdamped_rnn}
    \bm{v}(t + \dt) = (1 - \friction) \bvt + \underbrace{f(\brt, \but, \sigmaeta) - \brt}_{
    %\text{candidate } \Delta \widetilde{\br}(t)
    \drt{} \text{ if } \friction = 1
    },
    \\
    \drt{} = \bm{v}(t + \dt)
\end{multline}
\end{definition}
\paragraph{Biological plausibility.}
Several previous works in theoretical neuroscience have proposed mechanisms involving momentum, including Hamiltonian dynamics, for fast and improved sampling \cite{fast_sampling-based, natural_gradient_enables, biologically-plausible_mcmc,  neural_sampling_hierarchical, hamiltonian_brain, adaptation_accelerating}. Some have implemented these dynamics in E/I or spiking networks. Thus, it is conceivable that the brain could use momentum to quickly and efficiently sample \cite{fast_sampling-based}. Additionally, sampling schemes with momentum improve modeling of human random sequence generation \cite{explaining_the_flaws}, and momentum has been found in hippocampal replay trajectories \cite{large_majority}.
We propose momentum as a circuit mechanism for temporal compression in replay, but we cannot directly compare it to the physiological mechanisms behind temporal compression \textit{in vivo} because they are not yet fully understood.
Nonetheless, we identify two mechanisms for which momentum may be a reasonable initial approximation:
(1) Short-term facilitation can induce phase precession and temporal compression in spiking networks \cite{temporal_compression_mediated}. Extensions to rate networks are scant, so momentum may be a reasonable initial approximation.
(2) Short-term post-synaptic plasticity via NMDA receptors induces momentum-like effects in continuous attractor networks \cite{short-term_postsynaptic}.

%Moreover, our sampling mechanism shares conceptual ties with synaptic facilitation, a phenomenon associated with temporal compression \textit{in vivo} wherein inputs that successfully trigger output activity have a temporarily increased influence on output activity \citep{temporal_compression_mediated, phase_precession_through}; we leave a possible implementation of underdampening via synaptic facilitation to future work.
%Lastly, our mechanism could relate to momentum observed in replay trajectories \textit{in vivo} \citep{large_majority}.
\section{Numerical Results}

Now we examine how adaptation and underdampening (Equations \ref{eq:adaptation_definition} and \ref{eq:underdamped_rnn}) bias replay distributions in trained path-integrating RNNs.
We first see they counter each other: adaptation slows replay, while underdampening quickens it.
Then we confirm that adaptation induces exploration, and show that underdampening does not prevent exploration, but rather complements it by increasing path lengths. We note here that replay trajectories are obtained by first randomly initializing the hidden state of the RNN, following which we collect activations for several unrolled timesteps in the absence of inputs but in the presence of noise. Further details are provided in Appendix \ref{appendix:methods_expts}.
Here we use ReLU and leaky ReLU trained RNNs, as justified in \Cref{appendix:hyperparameters}, but in \Cref{appendix:tanh} we confirm that adaptation slows and underdampening compresses replay in tanh RNNs.

\paragraph{Experiments.}
We have five tasks on which we train RNNs and then examine how introducing sampling mechanisms post-training (adaptation strength $b_a$ and friction $\friction$) affects replay (quiescent activity) statistics.
We explain our implementation and many measured replay statistics in \Cref{appendix:methods}.
\begin{enumerate}
\item 1D Ornstein-Uhlenbeck (OU) process (\Cref{fig:1d_mean_std}): the optimal (with respect to \Cref{eq:loss_upper}) path-integrating RNN has a closed form (\Cref{eq:score_ou_main}), which we use in lieu of a trained RNN.
\item 2D T-maze and triangle (\Cref{fig:paths}): we simulate each direction (two in T-maze, six in triangle) as a 2D OU process, and train a ReLU RNN to integrate 2D paths from every direction.
\item Rat trajectories (\Cref{appendix:more_results}, \Cref{fig:paths_rat}): like \citet{sufficient_conditions}, we simulate 512 place cells that encode 2D directed (biased) or undirected (unbiased) random walks from RatInABox \citep{george2022ratinabox}, train a ReLU RNN to path-integrate place cell activity, and decode activity in 2D.
We use \textit{in silico} rat activity to mimic biological data because replay events \textit{in vivo} are sparse and difficult to extract from background noisy activity, therefore outside the scope of this work.
\end{enumerate}

\input{tikzfigs/paths_with_diagram}

%% ---------------------
%% Wasserstein distances
%% ---------------------
\begin{figure*}[h]
\centering
\begin{tikzpicture}

\pgfplotsset{/pgfplots/group/every plot/.append style = {
    ultra thick,%, no markers
    legend style={
        legend columns=2,
        anchor=south west}
}};

\begin{groupplot}[
    group style = {
        group size = 4 by 1,
        horizontal sep = 0.06\linewidth},
    %width = 0.27\linewidth,
    width = 0.2\linewidth,
    height = 0.15\linewidth,
    title style={yshift=-2.3mm,},
    axis x line*=bottom,
    axis y line*=left,    
    x dir=reverse,   
    xlabel=Friction $\friction$,
    xlabel shift={-1mm},
    ylabel style={rotate=-90},
    ylabel shift={-3mm},
    yticklabel style={/pgf/number format/.cd,
    fixed},
    grid,
    no marks,
    table/col sep=comma,
    cycle list name=mygradient
    ]

    \nextgroupplot[title=T-maze,
        ylabel=\column{\small Wasserstein\\ distance ($\downarrow$)},
        legend style={
            legend columns=3,
            at={(0.5,-1.5)},
            inner sep=1pt,
            font=\small}]
            %font=\footnotesize}]
    \foreach \bval in {0.0,0.5,1.0}{
        \addplot table[x=lambda_v, y={b_a=\bval}] {csv/wd/wd_tmaze.csv}; 
        \addlegendentryexpanded{ \bval};
    }

    \nextgroupplot[
        title=Triangle,
        title style={yshift=-0.5mm,}]
    \foreach \bval in {0.0,0.5,1.0}{
        \addplot table[x=lambda_v, y={b_a=\bval}] {csv/wd/wd_triangle.csv}; 
    }

    \nextgroupplot[
        title={Biased rat},
        width=0.25\linewidth,
        legend style={
            legend columns=4,
            at={(-0.1,-1.5)},
            row sep=-3pt,
            inner sep=1pt,
            font=\small}]
            %font=\footnotesize}]
    \foreach \bval in {0,1,3,5}{
        \addplot table[x=lambda_v, y expr=\thisrow{b_a=\bval}] {csv/wd/wds_biased.csv}; 
        \addlegendentryexpanded{ \bval};
    }

    \nextgroupplot[
        title={Unbiased rat},
        width=0.3\linewidth,
        xtick={1, 0.9, 0.8, 0.7},
        legend style={
            legend columns=4,
            at={(-0.1,-1.5)},
            row sep=-3pt,
            inner sep=1pt,
            font=\small}]
            %font=\footnotesize}]
    \foreach \bval in {0.0,0.1,0.2,0.3}{
        \addplot table[x=lambda_v, y expr=\thisrow{b_a=\bval}] {csv/wd/wds_unbiased.csv};
        \addlegendentryexpanded{ \bval};
    }
    %\addlegendentryexpanded{0};
    %\foreach \num in {1,2,3}{
    %    \addlegendentryexpanded{$\frac{\num}{10}$};
    %}

\end{groupplot} 

\node[inner sep=0] at (-1.4,-1.15) {\column{Adaptation\\strength $b_a$:}};

\end{tikzpicture}
\caption{
\textbf{Underdampening improves replay fidelity in the presence of adaptation.}
We compute the Wasserstein distance (dissimilarity) between awake and replay path distributions ($p( \timeset{\bst})$ and $p(\timeset{\brt})$), varying friction and adaptation strength (see \Cref{appendix:methods} for details).
While the two mechanisms both generally increase this distance, underdampening ($\friction < 1$) decreases it if adaptation is nonzero.
Like in \Cref{fig:paths}, underdampening counters adaptation-induced deviations.
}
\label{fig:wd}
\end{figure*}
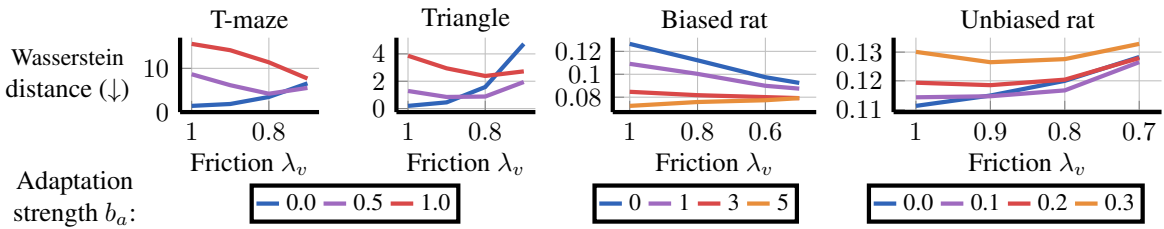

\paragraph{Underdampening and adaptation seem to counter each other.}
We initially observe that the two modifiers counteract each other: 
in Figures \ref{fig:1d_mean_std} and \ref{fig:paths}, adaptation repels trajectories away from attractors (such as endpoints), while underdampening accelerates trajectories towards them.
This makes sense given Remark \ref{remark:adaptation_negative_sign} and the nature of underdamped sampling.
Then we measure how the two modifiers affect the similarity between replay (internally-driven) and awake (observed) trajectories.
\Cref{fig:wd} shows they both generally decrease the similarity to awake trajectories, but they do counter each other insofar as underdampening, in the presence of adaptation, increases similarity to awake trajectories.
Thus far, underdampening counters adaptation qualitatively and statistically.

\newcommand{\rtMinTmaze}{-30}
\newcommand{\rtMaxTmazeHalf}{60}
\newcommand{\rtMaxTmaze}{120}
\newcommand{\rtMinTriangle}{-30}
\newcommand{\rtMaxTriangleHalf}{55}
\newcommand{\rtMaxTriangle}{110}
\newcommand{\rtMinBiased}{-60}
\newcommand{\rtMaxBiasedHalf}{150}
\newcommand{\rtMaxBiased}{300}
\pgfplotsset{
    colormap={bwrTmaze}{
        color(\rtMinTmaze)=(blue)
        color(0)=(white)
        color(\rtMaxTmaze)=(red)},
    colormap={bwrTriangle}{
        color(\rtMinTriangle)=(blue)
        color(0)=(white)
        color(\rtMaxTriangle)=(red)},
    colormap={bwrBiased}{
        color(\rtMinBiased)=(blue)
        color(0)=(white)
        color(\rtMaxBiased)=(red)}
}

%% ---------------------------------
%% reach time heatmaps (median only)
%% ---------------------------------
\begin{figure}%[htbp]
\centering
\begin{tikzpicture}

\begin{groupplot}[
    group style = {
        group size = 3 by 1,
        horizontal sep = 0.06\textwidth,
        vertical sep = 0.08\textwidth,
    },
    title style={yshift=-1.3ex,},
    width=0.38\linewidth,
    height=0.32\linewidth,
    % labels, ticks, and ticklabels
    xtick={0,1,2},
    xticklabels={$0$,$0.5$,$1$},
    xticklabel style={font=\small},
    %xlabel=\column{Adaptation\\strength $b_a$},
    xlabel style={yshift=1ex,},
    ytick={0,1,2,3},
    yticklabels={$1$,$0.9$,$0.8$,$0.7$},
    yticklabel style={font=\small},
    % ESSENTIAL params for correct rendering and avoiding unnecessary whitespace
    y dir=reverse,
    enlargelimits={abs=0.5},
    xmin=0, xmax=2,
    ymin=0, ymax=3,      
    ]

    %% -----------------
    %% Top row (medians)
    %% -----------------
    % T maze (median)
    \nextgroupplot[
        title=T-maze,
        point meta min=\rtMinTmaze,
        point meta max=\rtMaxTmaze,    
        colormap name=bwrTmaze,   
        % Bottom row plot + heatmap params
        ylabel=Friction $\friction$,
        ylabel style={yshift=-1mm},
        colorbar sampled, % ESSENTIAL FOR GROUPPLOTS
        colorbar horizontal,
        colorbar style={
            height=2mm,
            xtick={\rtMinTmaze, 0, \rtMaxTmazeHalf, \rtMaxTmaze},
            xticklabels={$\rtMinTmaze \quad \enspace$, $0$, $\rtMaxTmazeHalf$, $\rtMaxTmaze$},
            xticklabel style={font=\scriptsize},
            at={(0,-1.2)}
        }                
    ]
        \addplot [matrix plot*,point meta=explicit] file {csv/rt/med_rt_heatmap_tmaze.txt};

    % Triangle (median)
    \nextgroupplot[
        title=Triangle,
        title style={yshift=-0.8mm,},
        point meta min=\rtMinTriangle,
        point meta max=\rtMaxTriangle,
        xlabel={Adaptation strength $b_a$},
        colormap name=bwrTriangle,
        % Bottom row plot + heatmap params
        colorbar sampled, % ESSENTIAL FOR GROUPPLOTS
        colorbar horizontal,
        colorbar style={
            height=2mm,
            xtick={\rtMinTriangle, 0, \rtMaxTriangleHalf, \rtMaxTriangle},
            xticklabels={$\rtMinTriangle \quad \enspace$, $0$, $\rtMaxTriangleHalf$, $\rtMaxTriangle$},
            xticklabel style={font=\scriptsize},
            at={(0,-1.2)}
        }        
    ]
        \addplot [matrix plot*,point meta=explicit] file {csv/rt/med_rt_heatmap_triangle.txt};

    % Biased rat (median)
    \nextgroupplot[
        title=Biased rat,
        point meta min=\rtMinBiased,
        point meta max=\rtMaxBiased,
        colormap name=bwrBiased,
        % Biased rat heatmap params
        xmax=3,
        xtick={0,1,2,3},
        xticklabels={$0$, $1$, $3$, $5$},
        yticklabels={$1$, $0.8$, $0.6$, $0.5$},
        % Bottom row plot + heatmap params
        colorbar sampled, % ESSENTIAL FOR GROUPPLOTS
        colorbar horizontal,
        colorbar style={
            height=2mm,
            xtick={\rtMinBiased, 0, \rtMaxBiasedHalf, \rtMaxBiased},
            xticklabels={$\rtMinBiased \quad \enspace$, $0$, $\rtMaxBiasedHalf$, $\rtMaxBiased$},
            xticklabel style={font=\scriptsize},
            at={(0,-1.2)}
        }                
    ]
        \addplot [matrix plot*,point meta=explicit] file {csv/rt/med_rt_heatmap_biased.txt};

\end{groupplot}

%\node[inner sep=0] at (-1.6,-2) {\column{Change (\%)\\from awake\\ statistics:}};
\node[inner sep=0] at (3.5,-1.05) {Change (\%) from awake reach time:};

%\draw[thick] ($(cmap_cap.north west) + (0,0.1)$) -- ($(cmap_cap.north east) + (0,0.1)$);

\draw[thick] (-0.5,-1.85) rectangle (7,-0.8);

\node[inner sep=0] at (3, 1.7) {Median replay reach time $(\downarrow)$};

\end{tikzpicture}
\caption{
\textbf{Underdampening temporally compresses replay.}
We calculate how long it takes awake and replay trajectories to reach their endpoints.
Underdampening ($\friction < 1$) not only shortens this reach time, but makes it smaller than that of awake paths, \textit{temporally compressing} awake activity. 
See \Cref{appendix:more_results} \Cref{fig:reachtime_heatmaps_all} for mean reach times.
We do not include unbiased rat trajectories because they do not have defined endpoints, but we do confirm in \Cref{appendix:more_results} \Cref{fig:mean_displacement} that underdampening quickens them.
}
\label{fig:reachtime_heatmaps}
\end{figure}
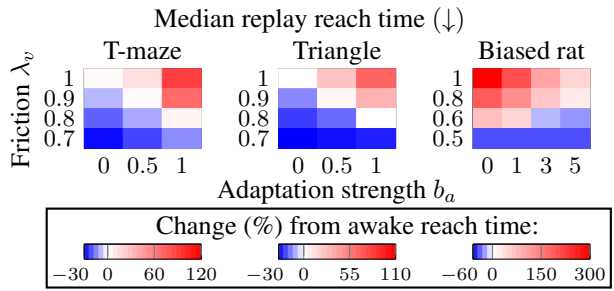

\paragraph{Adaptation slows replay, underdampening accelerates it.}
Next, we examine a key component of replay: speed.
In Figures \ref{fig:wd} and \ref{fig:reachtime_heatmaps}, we generally see that adaptation slows replay or increases the dissimilarity between replay and awake trajectories.
The exception is the biased rat trajectories, since they are the only task where all trajectories reach the same common goal.
Such path distributions resemble Ornstein-Uhlenbeck processes with steady-state mean $\mu = 0$: in such cases, adaptation actually accelerates convergence towards the steady-state.
In Figure \ref{fig:reachtime_heatmaps}, we see that underdampening increases replay speed, performing temporal compression relative to awake activity.

%% -------------------
%% Exploration metrics
%% -------------------
\begin{figure*}[h]
\centering
\begin{tikzpicture}

\pgfplotsset{/pgfplots/group/every plot/.append style = {
    ultra thick,%, no markers
    legend style={
        legend columns=2,
        anchor=south west}
}};

\begin{groupplot}[
    group style = {
        group size = 4 by 1,
        horizontal sep = 0.06\linewidth,
        vertical sep = 0.05\linewidth},
    %width = 0.27\linewith,
    width = 0.2\linewidth,
    height = 0.15\linewidth,
    title style={yshift=-2.3mm,},
    x dir=reverse,   
    axis x line*=bottom,
    axis y line*=left,
    xlabel={Friction $\friction$},
    xlabel shift={-1mm},    
    ylabel style={rotate=-90},
    ylabel shift={-3mm},
    grid,
    no marks,
    table/col sep=comma,
    cycle list name=mygradient
    ]

    % T-maze
    \nextgroupplot[
        title=T-maze,
        ylabel=\column{\small Mean path\\length ($\uparrow$)},
        legend style={
            legend columns=3,
            at={(0.5,-1.5)},
            inner sep=1pt,
            font=\small}]
    \foreach \bval in {0.0,0.5,1.0}{
        \addplot table[x=lambda_v, y={b_a=\bval}] {csv/exploration/exploration_tmaze_distances_avg.csv}; 
        \addlegendentryexpanded{ \bval};
    }

    % Triangle
    \nextgroupplot[
        title=Triangle,
        title style={yshift=-0.5mm,},]
    \foreach \bval in {0.0,0.5,1.0}{
        \addplot table[x=lambda_v, y={b_a=\bval}] {csv/exploration/exploration_triangle_distances_avg.csv};  
    }

    % Biased rat
    \nextgroupplot[
        title=Biased rat,
        width=0.25\linewidth,
        legend style={
            at={(0,-1.5)},
            legend columns=4,
            row sep=-3pt,
            inner sep=1pt,
            font=\small}]
    \foreach \bval in {0,1,3,5}{
        \addplot table[x=lambda_v, y={b_a=\bval}] {csv/exploration/exploration_distance_biased.csv};  
        \addlegendentryexpanded{ \bval};
    }    

    % Unbiased rat
    \nextgroupplot[
        title=Unbiased rat,
        width=0.3\linewidth,
        legend style={
            at={(0.1,-1.5)},
            legend columns=4,
            row sep=-3pt,
            inner sep=1pt,
            font=\small}]
    \foreach \bval in {0.0,0.1,0.2,0.3}{
        \addplot table[x=lambda_v, y={b_a=\bval}] {csv/exploration/exploration_distance_unbiased.csv};  
        \addlegendentryexpanded{ \bval};
    }

\end{groupplot} 

\node[inner sep=0] at (-1.3,-1.15) {\column{Adaptation\\strength $b_a$:}};

\end{tikzpicture}
\caption{
\textbf{On average, underdampening increases exploration via path length.}
In here and \Cref{fig:region_count_avg} we simulate replay paths for more time than awake paths, varying friction and adaptation strength.
Shown above, underdampening ($\friction < 1$) increases the average length of replay paths.
It may do so via increased replay speeds (\Cref{appendix:more_results} \Cref{fig:mean_displacement}) or via additional path transitions (\Cref{fig:region_count_avg}); the latter is how adaptation increases path length.
}
\label{fig:exploration_metrics}
\end{figure*}
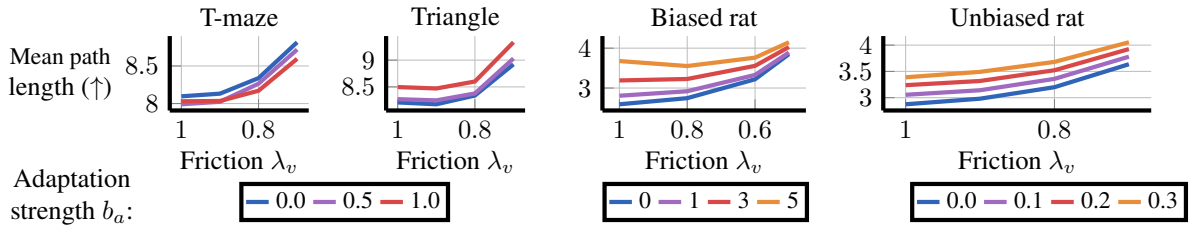

\input{tikzfigs/exploration_panels}

\paragraph{Underdampening complements exploration from adaptation.}
We have established that underdampening counters adaptation, both qualitatively and in terms of speed.
However, we do not want to merely propose a mechanism that undoes adaptation.
Adaptation can induce exploration, i.e., prevent replay trajectories from getting stuck in attractors.
We find that underdampening does not on average prevent adaptation-induced exploration.
In fact, underdampening complements adaptation for exploration: underdamped paths travel farther (Figures \ref{fig:exploration_metrics} and \ref{fig:region_count_avg}), have more variance (\Cref{appendix:more_results} \Cref{fig:time_variances}), and generally exhibit the same, if not more, exploratory behavior as they did with only adaptation (\Cref{fig:region_count_avg}).
Underdampening maintains exploration while counteracting the slowness from adaptation.
%, thus being a promising tool for future studies of replay.
\section{Conclusions and Future Work}

We have re-applied Langevin sampling theory to replay in sequential predictive learning networks, producing theoretically and confirming empirically three key insights:
(1) estimating the per-timestep score function of RNN activity is challenging, but does benefit from linear leakage;
(2) adaptation (negative feedback) is a variant of underdamped Langevin sampling that encourages exploration (as shown in prior works) but also slows replay;
(3) our new underdampening mechanism (momentum) temporally compresses replay while also increasing exploration from adaptation.
These findings improve our understanding of biological neural networks that produce replay, like the hippocampus.
Our proposed underdampening mechanism via momentum could be tied to short-term facilitation (which can be probed experimentally), connected to specific subregions in the hippocampus (like \citet{predictive_sequence_learning}), or refined through insights from existing RNNs with momentum \citep{momentumrnn}.
Future efforts could confirm our findings in more complex environments \cite{honeycomb_maze, sequential_predictive_learning}.
Like preceding works, our network models are rate-based, but extending our work to spike-based models of replay and sequential predictive learning \citep{sequence_anticipation, predictive_learning_rules, learning_predictive_cognitive} would be interesting (i.e., extending Langevin sampling to Poisson processes).
Several avenues remain for connecting Langevin sampling to neural replay.

%\section{Acknowledgements}
%This work was supported by NSF grants CCF-1911094 and IIS-1730574; ONR grants N00014-23-1-2714, N00014-24-1-2225, and MURI N00014-20-1-2787; AFOSR grant FA9550-22-1-0060; DOE grant DE-SC0020345; DOI grant 140D0423C0076; and a Vannevar Bush Faculty Fellowship, ONR grant N00014-18-1-2047.
% TODO: acknowledge Ali, Colin, Tan, Kemere, and the 3 RIKEN ppl

\section*{Impact Statement}
This work aims to advance our understanding of the computations underlying replay in the brain. It is a theoretical study that aims to advance our understanding of neuroscience through the use of artificial neural network models, as is common in computational neuroscience. There may be many potential societal consequences of our work in the long-term, none of which we feel must be specifically highlighted here.

\section*{Reproducibility Statement}
For our theoretical contributions, i.e., Theorems 3.2 and 4.3, we provide proofs in Appendix \ref{sec:appendix_score} and \ref{sec:appendix_adaptation_sde} (proofs for Theorems 2.7 and 2.8 may be found in Section 2 of \citet{sufficient_conditions}). Details on our numerical experiments, including associated hyperparameters and implementation details are provided in Appendix \ref{appendix:methods}. Our code has been submitted as Supplementary Material (see \texttt{README.md} for instructions) and will be made publicly available upon publication.

\bibliographystyle{icml2026}
\bibliography{_bib,_bib_nanda}

\begin{thebibliography}{102}
\providecommand{\natexlab}[1]{#1}
\providecommand{\url}[1]{\texttt{#1}}
\expandafter\ifx\csname urlstyle\endcsname\relax
  \providecommand{\doi}[1]{doi: #1}\else
  \providecommand{\doi}{doi: \begingroup \urlstyle{rm}\Url}\fi

\bibitem[Aitchison \& Lengyel(2016)Aitchison and Lengyel]{hamiltonian_brain}
Aitchison, L. and Lengyel, M.
\newblock The {Hamiltonian} brain: Efficient probabilistic inference with excitatory-inhibitory neural circuit dynamics.
\newblock \emph{PLoS Computational biology}, 12\penalty0 (12), 2016.

\bibitem[Aitken \& Kok(2022)Aitken and Kok]{hippocampal_representations_switch}
Aitken, F. and Kok, P.
\newblock Hippocampal representations switch from errors to predictions during acquisition of predictive associations.
\newblock \emph{Nature Communications}, 13\penalty0 (1), 2022.

\bibitem[Alemohammad et~al.(2024)Alemohammad, Casco-Rodriguez, Luzi, Humayun, Babaei, LeJeune, Siahkoohi, and Baraniuk]{self_consuming}
Alemohammad, S., Casco-Rodriguez, J., Luzi, L., Humayun, A.~I., Babaei, H., LeJeune, D., Siahkoohi, A., and Baraniuk, R.~G.
\newblock Self-consuming generative models go {MAD}.
\newblock In \emph{International Conference on Learning Representations (ICLR)}, 2024.

\bibitem[Ali et~al.(2022)Ali, Ahmad, de~Groot, van Gerven, and Kietzmann]{predictive_coding_consequence}
Ali, A., Ahmad, N., de~Groot, E., van Gerven, M. A.~J., and Kietzmann, T.~C.
\newblock Predictive coding is a consequence of energy efficiency in recurrent neural networks.
\newblock \emph{Patterns}, 3\penalty0 (12), 2022.

\bibitem[Asabuki \& Fukai(2025)Asabuki and Fukai]{predictive_learning_rules}
Asabuki, T. and Fukai, T.
\newblock Predictive learning rules generate a cortical-like replay of probabilistic sensory experiences.
\newblock \emph{eLife}, 13, 2025.

\bibitem[Azizi et~al.(2013)Azizi, Wiskott, and Cheng]{computational_model_preplay}
Azizi, A.~H., Wiskott, L., and Cheng, S.
\newblock A computational model for preplay in the hippocampus.
\newblock \emph{Frontiers in computational neuroscience}, 7:\penalty0 161, 2013.

\bibitem[Battaglia \& Treves(1998)Battaglia and Treves]{attractor_neural_networks_storing}
Battaglia, F.~P. and Treves, A.
\newblock Attractor neural networks storing multiple space representations: a model for hippocampal place fields.
\newblock \emph{Physical Review E}, 58\penalty0 (6), 1998.

\bibitem[Benda(2021)]{neural_adaptation}
Benda, J.
\newblock Neural adaptation.
\newblock \emph{Current Biology}, 31\penalty0 (3), 2021.

\bibitem[Besag(1994)]{besag1994}
Besag, J.~E.
\newblock Discussion of ``{R}epresentations of knowledge in complex systems'' by {Ulf Grenander} and {Michael I Miller}.
\newblock \emph{Journal of the Royal Statistics Society B}, 56\penalty0 (4):\penalty0 591--592, 1994.

\bibitem[Bonneel et~al.(2015)Bonneel, Rabin, Peyr{\'e}, and Pfister]{sliced_radon_wasserstein}
Bonneel, N., Rabin, J., Peyr{\'e}, G., and Pfister, H.
\newblock Sliced and radon wasserstein barycenters of measures.
\newblock \emph{Journal of Mathematical Imaging and Vision}, 51\penalty0 (1), 2015.

\bibitem[Bono et~al.(2023)Bono, Zannone, Pedrosa, and Clopath]{learning_predictive_cognitive}
Bono, J., Zannone, S., Pedrosa, V., and Clopath, C.
\newblock Learning predictive cognitive maps with spiking neurons during behavior and replays.
\newblock \emph{Elife}, 12, 2023.

\bibitem[Burgess et~al.(1994)Burgess, Recce, and O'Keefe]{model_hippocampal}
Burgess, N., Recce, M., and O'Keefe, J.
\newblock A model of hippocampal function.
\newblock \emph{Neural networks}, 7\penalty0 (6-7), 1994.

\bibitem[Buzs{\'a}ki(1986)]{hippocampal_sharp_waves}
Buzs{\'a}ki, G.
\newblock Hippocampal sharp waves: their origin and significance.
\newblock \emph{Brain research}, 398\penalty0 (2), 1986.

\bibitem[Buzs{\'a}ki(1989)]{two-stage_model}
Buzs{\'a}ki, G.
\newblock Two-stage model of memory trace formation: a role for “noisy” brain states.
\newblock \emph{Neuroscience}, 31\penalty0 (3), 1989.

\bibitem[Buzs{\'a}ki(2015)]{hippocampal_sharp_wave_ripple}
Buzs{\'a}ki, G.
\newblock Hippocampal sharp wave-ripple: A cognitive biomarker for episodic memory and planning.
\newblock \emph{Hippocampus}, 25\penalty0 (10), 2015.

\bibitem[Castillo et~al.(2024)Castillo, Le{\'o}n-Villagr{\'a}, Chater, and Sanborn]{explaining_the_flaws}
Castillo, L., Le{\'o}n-Villagr{\'a}, P., Chater, N., and Sanborn, A.
\newblock Explaining the flaws in human random generation as local sampling with momentum.
\newblock \emph{PLOS Computational Biology}, 20\penalty0 (1), 2024.

\bibitem[Chen et~al.(2024)Chen, Zhang, Cameron, and Sejnowski]{predictive_sequence_learning}
Chen, Y., Zhang, H., Cameron, M., and Sejnowski, T.
\newblock Predictive sequence learning in the hippocampal formation.
\newblock \emph{Neuron}, 112\penalty0 (15), 2024.

\bibitem[Cheng et~al.(2018)Cheng, Chatterji, Bartlett, and Jordan]{underdamped_langevin_mcmc}
Cheng, X., Chatterji, N.~S., Bartlett, P.~L., and Jordan, M.~I.
\newblock Underdamped {Langevin MCMC}: A non-asymptotic analysis.
\newblock In \emph{Proceedings of the 31st Conference On Learning Theory}, volume~75, 2018.

\bibitem[Chettih et~al.(2024)Chettih, Mackevicius, Hale, and Aronov]{barcoding_episodic}
Chettih, S.~N., Mackevicius, E.~L., Hale, S., and Aronov, D.
\newblock Barcoding of episodic memories in the hippocampus of a food-caching bird.
\newblock \emph{Cell}, 187\penalty0 (8), 2024.

\bibitem[Chrastil(2025)]{human_path_integration}
Chrastil, E.~R.
\newblock Human path integration and the neural underpinnings.
\newblock In \emph{Encyclopedia of the Human Brain (Second Edition)}, pp.\  157--170. Elsevier, 2025.
\newblock ISBN 978-0-12-820481-8.
\newblock \doi{https://doi.org/10.1016/B978-0-12-820480-1.00016-4}.

\bibitem[Chu et~al.(2024)Chu, Ji, Zuo, Mi, Zhang, Huang, Bush, Burgess, and Wu]{chu2023firing}
Chu, T., Ji, Z., Zuo, J., Mi, Y., Zhang, W., Huang, T., Bush, D., Burgess, N., and Wu, S.
\newblock Firing rate adaptation affords place cell theta sweeps, phase precession and procession.
\newblock \emph{bioRxiv}, 2024.

\bibitem[Churchland \& Shenoy(2024)Churchland and Shenoy]{preparatory_activity}
Churchland, M.~M. and Shenoy, K.~V.
\newblock Preparatory activity and the expansive null-space.
\newblock \emph{Nature Reviews Neuroscience}, 25\penalty0 (4), 2024.

\bibitem[Croitoru et~al.(2023)Croitoru, Hondru, Ionescu, and Shah]{diffusion_models_in_vision}
Croitoru, F.-A., Hondru, V., Ionescu, R.~T., and Shah, M.
\newblock Diffusion models in vision: A survey.
\newblock \emph{IEEE Transactions on Pattern Analysis and Machine Intelligence}, 45\penalty0 (9), 2023.

\bibitem[Cueva \& Wei(2018)Cueva and Wei]{cueva2018emergence}
Cueva, C.~J. and Wei, X.-X.
\newblock Emergence of grid-like representations by training recurrent neural networks to perform spatial localization.
\newblock In \emph{International Conference on Learning Representations}, 2018.

\bibitem[Cueva et~al.(2020)Cueva, Wang, Chin, and Wei]{Cueva2020Emergence}
Cueva, C.~J., Wang, P.~Y., Chin, M., and Wei, X.-X.
\newblock Emergence of functional and structural properties of the head direction system by optimization of recurrent neural networks.
\newblock In \emph{International Conference on Learning Representations}, 2020.

\bibitem[Davachi \& DuBrow(2015)Davachi and DuBrow]{hippocampus_preserves_order}
Davachi, L. and DuBrow, S.
\newblock How the hippocampus preserves order: the role of prediction and context.
\newblock \emph{Trends in cognitive sciences}, 19\penalty0 (2), 2015.

\bibitem[Davidson et~al.(2009)Davidson, Kloosterman, and Wilson]{hippocampal_replay_extended}
Davidson, T.~J., Kloosterman, F., and Wilson, M.~A.
\newblock Hippocampal replay of extended experience.
\newblock \emph{Neuron}, 63\penalty0 (4), 2009.

\bibitem[Deperrois et~al.(2022)Deperrois, Petrovici, Senn, and Jordan]{deperrois2022learning}
Deperrois, N., Petrovici, M.~A., Senn, W., and Jordan, J.
\newblock Learning cortical representations through perturbed and adversarial dreaming.
\newblock \emph{eLife}, 11:\penalty0 e76384, 2022.
\newblock ISSN 2050-084X.

\bibitem[Dong \& Wu(2023)Dong and Wu]{neural_sampling_hierarchical}
Dong, X. and Wu, S.
\newblock Neural sampling in hierarchical exponential-family energy-based models.
\newblock \emph{Advances in Neural Information Processing Systems (NeurIPS)}, 36, 2023.

\bibitem[Dong et~al.(2021)Dong, Chu, Huang, Ji, and Wu]{noisy_adaptation}
Dong, X., Chu, T., Huang, T., Ji, Z., and Wu, S.
\newblock Noisy adaptation generates {L\'evy} flights in attractor neural networks.
\newblock \emph{Advances in Neural Information Processing Systems (NeurIPS)}, 34, 2021.

\bibitem[Dong et~al.(2022)Dong, Ji, Chu, Huang, Zhang, and Wu]{adaptation_accelerating}
Dong, X., Ji, Z., Chu, T., Huang, T., Zhang, W., and Wu, S.
\newblock Adaptation accelerating sampling-based bayesian inference in attractor neural networks.
\newblock \emph{Advances in Neural Information Processing Systems (NeurIPS)}, 35, 2022.

\bibitem[Euston et~al.(2007)Euston, Tatsuno, and McNaughton]{euston2007fast}
Euston, D.~R., Tatsuno, M., and McNaughton, B.~L.
\newblock Fast-forward playback of recent memory sequences in prefrontal cortex during sleep.
\newblock \emph{Science}, 318\penalty0 (5853):\penalty0 1147--1150, 2007.

\bibitem[Farooq \& Dragoi(2019)Farooq and Dragoi]{emergence_preconfigured}
Farooq, U. and Dragoi, G.
\newblock Emergence of preconfigured and plastic time-compressed sequences in early postnatal development.
\newblock \emph{Science}, 363\penalty0 (6423), 2019.

\bibitem[Foster(2017)]{foster2017replay}
Foster, D.~J.
\newblock Replay comes of age.
\newblock \emph{Annual Review of Neuroscience}, 40:\penalty0 581--602, 2017.
\newblock ISSN 1545-4126.

\bibitem[Furlong et~al.(2024)Furlong, Simone, Dumont, Bartlett, Stewart, Orchard, and Eliasmith]{biologically-plausible_mcmc}
Furlong, P.~M., Simone, K., Dumont, N. S.-Y., Bartlett, M., Stewart, T.~C., Orchard, J., and Eliasmith, C.
\newblock Biologically-plausible {Markov Chain Monte Carlo} sampling from vector symbolic algebra-encoded distributions.
\newblock In \emph{International Conference on Artificial Neural Networks}, 2024.

\bibitem[Gardner et~al.(2022)Gardner, Hermansen, Pachitariu, Burak, Baas, Dunn, Moser, and Moser]{gardner2022toroidal}
Gardner, R.~J., Hermansen, E., Pachitariu, M., Burak, Y., Baas, N.~A., Dunn, B.~A., Moser, M.-B., and Moser, E.~I.
\newblock Toroidal topology of population activity in grid cells.
\newblock \emph{Nature}, 602\penalty0 (7895):\penalty0 123–128, 2022.
\newblock ISSN 1476-4687.

\bibitem[George et~al.(2024)George, Rastogi, de~Cothi, Clopath, Stachenfeld, and Barry]{george2022ratinabox}
George, T.~M., Rastogi, M., de~Cothi, W., Clopath, C., Stachenfeld, K., and Barry, C.
\newblock {RatInABox}, a toolkit for modelling locomotion and neuronal activity in continuous environments.
\newblock \emph{eLife}, 13:\penalty0 e85274, 2024.
\newblock ISSN 2050-084X.

\bibitem[Gozalo-Brizuela \& Garrido-Merchan(2023)Gozalo-Brizuela and Garrido-Merchan]{chatgpt_is_not}
Gozalo-Brizuela, R. and Garrido-Merchan, E.~C.
\newblock {ChatGPT} is not all you need. a state of the art review of large generative {AI} models.
\newblock \emph{arXiv preprint arXiv:2301.04655}, 2023.

\bibitem[Gutkin \& Zeldenrust(2014)Gutkin and Zeldenrust]{spike_frequency_adaptation}
Gutkin, B. and Zeldenrust, F.
\newblock Spike frequency adaptation.
\newblock \emph{Scholarpedia}, 9\penalty0 (2), 2014.

\bibitem[Haga \& Fukai(2018)Haga and Fukai]{haga2018recurrent}
Haga, T. and Fukai, T.
\newblock Recurrent network model for learning goal-directed sequences through reverse replay.
\newblock \emph{eLife}, 7:\penalty0 e34171, 2018.
\newblock ISSN 2050-084X.

\bibitem[Hennequin et~al.(2014)Hennequin, Aitchison, and Lengyel]{fast_sampling-based}
Hennequin, G., Aitchison, L., and Lengyel, M.
\newblock Fast sampling-based inference in balanced neuronal networks.
\newblock \emph{Advances in Neural Information Processing Systems (NeurIPS)}, 27, 2014.

\bibitem[Hoffman \& McNaughton(2002)Hoffman and McNaughton]{hoffman2002coordinated}
Hoffman, K.~L. and McNaughton, B.~L.
\newblock Coordinated reactivation of distributed memory traces in primate neocortex.
\newblock \emph{Science}, 297\penalty0 (5589):\penalty0 2070--2073, 2002.

\bibitem[Hopfield(2010)]{hopfield2009neurodynamics}
Hopfield, J.~J.
\newblock Neurodynamics of mental exploration.
\newblock \emph{Proceedings of the National Academy of Sciences}, 107\penalty0 (4):\penalty0 1648--1653, 2010.

\bibitem[Itskov et~al.(2011)Itskov, Curto, Pastalkova, and Buzs{\'a}ki]{cell_assembly_sequences}
Itskov, V., Curto, C., Pastalkova, E., and Buzs{\'a}ki, G.
\newblock Cell assembly sequences arising from spike threshold adaptation keep track of time in the hippocampus.
\newblock \emph{Journal of Neuroscience}, 31\penalty0 (8), 2011.

\bibitem[Jaramillo \& Kempter(2017)Jaramillo and Kempter]{phase_precession_neural_code}
Jaramillo, J. and Kempter, R.
\newblock Phase precession: a neural code underlying episodic memory?
\newblock \emph{Current Opinion in Neurobiology}, 43, 2017.

\bibitem[Johnson \& Redish(2007)Johnson and Redish]{neural_ensembles_ca3}
Johnson, A. and Redish, A.~D.
\newblock Neural ensembles in {CA3} transiently encode paths forward of the animal at a decision point.
\newblock \emph{Journal of Neuroscience}, 27\penalty0 (45), 2007.

\bibitem[Kang \& DeWeese(2019)Kang and DeWeese]{kang2019replay}
Kang, L. and DeWeese, M.~R.
\newblock Replay as wavefronts and theta sequences as bump oscillations in a grid cell attractor network.
\newblock \emph{eLife}, 8:\penalty0 e46351, 2019.
\newblock ISSN 2050-084X.

\bibitem[Kay et~al.(2020)Kay, Chung, Sosa, Schor, Karlsson, Larkin, Liu, and Frank]{constant_sub-second}
Kay, K., Chung, J.~E., Sosa, M., Schor, J.~S., Karlsson, M.~P., Larkin, M.~C., Liu, D.~F., and Frank, L.~M.
\newblock Constant sub-second cycling between representations of possible futures in the hippocampus.
\newblock \emph{Cell}, 180\penalty0 (3), 2020.

\bibitem[Kenet et~al.(2003)Kenet, Bibitchkov, Tsodyks, Grinvald, and Arieli]{kenet2003spontaneously}
Kenet, T., Bibitchkov, D., Tsodyks, M., Grinvald, A., and Arieli, A.
\newblock Spontaneously emerging cortical representations of visual attributes.
\newblock \emph{Nature}, 425\penalty0 (6961):\penalty0 954–956, 2003.
\newblock ISSN 1476-4687.

\bibitem[Krause \& Drugowitsch(2022)Krause and Drugowitsch]{large_majority}
Krause, E.~L. and Drugowitsch, J.
\newblock A large majority of awake hippocampal sharp-wave ripples feature spatial trajectories with momentum.
\newblock \emph{Neuron}, 110\penalty0 (4), 2022.

\bibitem[Krishna et~al.(2024)Krishna, Bredenberg, Levenstein, Richards, and Lajoie]{sufficient_conditions}
Krishna, N.~H., Bredenberg, C., Levenstein, D., Richards, B.~A., and Lajoie, G.
\newblock Sufficient conditions for offline reactivation in recurrent neural networks.
\newblock In \emph{International Conference on Learning Representations (ICLR)}, 2024.

\bibitem[Lee \& Wilson(2002)Lee and Wilson]{lee2002memory}
Lee, A.~K. and Wilson, M.~A.
\newblock Memory of sequential experience in the hippocampus during slow wave sleep.
\newblock \emph{Neuron}, 36\penalty0 (6):\penalty0 1183–1194, 2002.
\newblock ISSN 0896-6273.

\bibitem[Leibold et~al.(2008)Leibold, Gundlfinger, Schmidt, Thurley, Schmitz, and Kempter]{temporal_compression_mediated}
Leibold, C., Gundlfinger, A., Schmidt, R., Thurley, K., Schmitz, D., and Kempter, R.
\newblock Temporal compression mediated by short-term synaptic plasticity.
\newblock \emph{Proceedings of the National Academy of Sciences}, 105\penalty0 (11), 2008.

\bibitem[Levenstein et~al.(2019)Levenstein, Buzs{\'a}ki, and Rinzel]{nrem_sleep}
Levenstein, D., Buzs{\'a}ki, G., and Rinzel, J.
\newblock {NREM} sleep in the rodent neocortex and hippocampus reflects excitable dynamics.
\newblock \emph{Nature Communications}, 10\penalty0 (1), 2019.

\bibitem[Levenstein et~al.(2024)Levenstein, Efremov, Eyono, Peyrache, and Richards]{sequential_predictive_learning}
Levenstein, D., Efremov, A., Eyono, R.~H., Peyrache, A., and Richards, B.
\newblock Sequential predictive learning is a unifying theory for hippocampal representation and replay.
\newblock \emph{bioRxiv}, 2024.

\bibitem[Levy(1989)]{computational_approach}
Levy, W.~B.
\newblock A computational approach to hippocampal function.
\newblock In \emph{Psychology of learning and motivation}, volume~23. Elsevier, 1989.

\bibitem[Li et~al.(2022)Li, Thickstun, Gulrajani, Liang, and Hashimoto]{diffusion-lm}
Li, X., Thickstun, J., Gulrajani, I., Liang, P.~S., and Hashimoto, T.~B.
\newblock Diffusion-{LM} improves controllable text generation.
\newblock \emph{Advances in Neural Information Processing Systems (NeurIPS)}, 35, 2022.

\bibitem[Li et~al.(2024)Li, Chu, and Wu]{dynamics_of_adaptive_canns}
Li, Y., Chu, T., and Wu, S.
\newblock Dynamics of adaptive continuous attractor neural networks.
\newblock \emph{arXiv preprint arXiv:2410.06517}, 2024.

\bibitem[Litwin-Kumar \& Doiron(2014)Litwin-Kumar and Doiron]{litwin2014formation}
Litwin-Kumar, A. and Doiron, B.
\newblock Formation and maintenance of neuronal assemblies through synaptic plasticity.
\newblock \emph{Nature Communications}, 5\penalty0 (1):\penalty0 5319, 2014.
\newblock ISSN 2041-1723.

\bibitem[Luzi et~al.(2024)Luzi, Mayer, Casco-Rodriguez, Siahkoohi, and Baraniuk]{boomerang}
Luzi, L., Mayer, P.~M., Casco-Rodriguez, J., Siahkoohi, A., and Baraniuk, R.
\newblock Boomerang: Local sampling on image manifolds using diffusion models.
\newblock \emph{Transactions on Machine Learning Research (TMLR)}, 2024.

\bibitem[Masset et~al.(2022)Masset, Zavatone-Veth, Connor, Murthy, and Pehlevan]{natural_gradient_enables}
Masset, P., Zavatone-Veth, J., Connor, J.~P., Murthy, V., and Pehlevan, C.
\newblock Natural gradient enables fast sampling in spiking neural networks.
\newblock \emph{Advances in Neural Information Processing Systems (NeurIPS)}, 35, 2022.

\bibitem[McNamee et~al.(2021)McNamee, Stachenfeld, Botvinick, and Gershman]{flexible_modulation}
McNamee, D.~C., Stachenfeld, K.~L., Botvinick, M.~M., and Gershman, S.~J.
\newblock Flexible modulation of sequence generation in the entorhinal--hippocampal system.
\newblock \emph{Nature Neuroscience}, 24\penalty0 (6), 2021.

\bibitem[McNaughton et~al.(1996)McNaughton, Barnes, Gerrard, Gothard, Jung, Knierim, Kudrimoti, Qin, Skaggs, Suster, et~al.]{deciphering_hippocampal}
McNaughton, B.~L., Barnes, C.~A., Gerrard, J.~L., Gothard, K., Jung, M.~W., Knierim, J.~J., Kudrimoti, H., Qin, Y., Skaggs, W., Suster, M., et~al.
\newblock Deciphering the hippocampal polyglot: the hippocampus as a path integration system.
\newblock \emph{Journal of Experimental Biology}, 199\penalty0 (1), 1996.

\bibitem[Michelmann et~al.(2019)Michelmann, Staresina, Bowman, and Hanslmayr]{speed_time_compressed}
Michelmann, S., Staresina, B.~P., Bowman, H., and Hanslmayr, S.
\newblock Speed of time-compressed forward replay flexibly changes in human episodic memory.
\newblock \emph{Nature Human Behaviour}, 3\penalty0 (2), 2019.

\bibitem[Miller et~al.(2023)Miller, Jacob, Ramsaran, De~Snoo, Josselyn, and Frankland]{emergence_predictive_model}
Miller, A.~M., Jacob, A.~D., Ramsaran, A.~I., De~Snoo, M.~L., Josselyn, S.~A., and Frankland, P.~W.
\newblock Emergence of a predictive model in the hippocampus.
\newblock \emph{Neuron}, 111\penalty0 (12), 2023.

\bibitem[Miller et~al.(2017)Miller, Botvinick, and Brody]{dorsal_hippocampus}
Miller, K.~J., Botvinick, M.~M., and Brody, C.~D.
\newblock Dorsal hippocampus contributes to model-based planning.
\newblock \emph{Nature neuroscience}, 20\penalty0 (9), 2017.

\bibitem[Milstein et~al.(2023)Milstein, Tran, Ng, and Soltesz]{milstein2023offline}
Milstein, A.~D., Tran, S., Ng, G., and Soltesz, I.
\newblock Offline memory replay in recurrent neuronal networks emerges from constraints on online dynamics.
\newblock \emph{The Journal of Physiology}, 601\penalty0 (15):\penalty0 3241--3264, 2023.

\bibitem[Miyasawa(1961)]{miyasawa1961empirical}
Miyasawa, K.
\newblock An empirical {B}ayes estimator of the mean of a normal population.
\newblock \emph{Bulletin of the International Statistical Institute}, 38\penalty0 (4):\penalty0 181--188, 1961.

\bibitem[N{\'a}dasdy et~al.(1999)N{\'a}dasdy, Hirase, Czurk{\'o}, Csicsvari, and Buzs{\'a}ki]{nadasdy1999replay}
N{\'a}dasdy, Z., Hirase, H., Czurk{\'o}, A., Csicsvari, J., and Buzs{\'a}ki, G.
\newblock Replay and time compression of recurring spike sequences in the hippocampus.
\newblock \emph{Journal of Neuroscience}, 19\penalty0 (21):\penalty0 9497--9507, 1999.
\newblock ISSN 0270-6474.

\bibitem[Nguyen et~al.(2020)Nguyen, Baraniuk, Bertozzi, Osher, and Wang]{momentumrnn}
Nguyen, T., Baraniuk, R., Bertozzi, A., Osher, S., and Wang, B.
\newblock {MomentumRNN}: Integrating momentum into recurrent neural networks.
\newblock \emph{Advances in Neural Information Processing Systems (NeurIPS)}, 33, 2020.

\bibitem[O'Keefe \& Nadel(1978)O'Keefe and Nadel]{hippocampus_cognitive_map}
O'Keefe, J. and Nadel, L.
\newblock \emph{The hippocampus as a cognitive map}.
\newblock Oxford University Press, 1978.

\bibitem[Panaretos \& Zemel(2019)Panaretos and Zemel]{statistical_aspects_wasserstein}
Panaretos, V.~M. and Zemel, Y.
\newblock Statistical aspects of wasserstein distances.
\newblock \emph{Annual Review of Statistics and its Application}, 6\penalty0 (1), 2019.

\bibitem[Pavliotis(2014)]{stochastic_processes}
Pavliotis, G.~A.
\newblock Stochastic processes and applications.
\newblock \emph{Texts in Applied Mathematics}, 60, 2014.

\bibitem[Peyrache et~al.(2009)Peyrache, Khamassi, Benchenane, Wiener, and Battaglia]{peyrache2009replay}
Peyrache, A., Khamassi, M., Benchenane, K., Wiener, S.~I., and Battaglia, F.~P.
\newblock Replay of rule-learning related neural patterns in the prefrontal cortex during sleep.
\newblock \emph{Nature Neuroscience}, 12\penalty0 (7):\penalty0 919–926, 2009.
\newblock ISSN 1546-1726.

\bibitem[Peyrache et~al.(2015)Peyrache, Lacroix, Petersen, and Buzs{\'a}ki]{peyrache2015internally}
Peyrache, A., Lacroix, M.~M., Petersen, P.~C., and Buzs{\'a}ki, G.
\newblock Internally organized mechanisms of the head direction sense.
\newblock \emph{Nature Neuroscience}, 18\penalty0 (4):\penalty0 569–575, 2015.
\newblock ISSN 1546-1726.

\bibitem[Pfeiffer(2020)]{content_of_hippocampal}
Pfeiffer, B.~E.
\newblock The content of hippocampal “replay”.
\newblock \emph{Hippocampus}, 30\penalty0 (1), 2020.

\bibitem[Pfeiffer \& Foster(2013)Pfeiffer and Foster]{hippocampal_place-cell_sequences}
Pfeiffer, B.~E. and Foster, D.~J.
\newblock Hippocampal place-cell sequences depict future paths to remembered goals.
\newblock \emph{Nature}, 497\penalty0 (7447), 2013.

\bibitem[Recanatesi et~al.(2021)Recanatesi, Farrell, Lajoie, Deneve, Rigotti, and Shea-Brown]{predictive_learning_network_mechanism}
Recanatesi, S., Farrell, M., Lajoie, G., Deneve, S., Rigotti, M., and Shea-Brown, E.
\newblock Predictive learning as a network mechanism for extracting low-dimensional latent space representations.
\newblock \emph{Nature Communications}, 12\penalty0 (1), 2021.

\bibitem[Samsonovich \& McNaughton(1997)Samsonovich and McNaughton]{path_integration}
Samsonovich, A. and McNaughton, B.~L.
\newblock Path integration and cognitive mapping in a continuous attractor neural network model.
\newblock \emph{Journal of Neuroscience}, 17\penalty0 (15), 1997.

\bibitem[Saponati \& Vinck(2023)Saponati and Vinck]{sequence_anticipation}
Saponati, M. and Vinck, M.
\newblock Sequence anticipation and spike-timing-dependent plasticity emerge from a predictive learning rule.
\newblock \emph{Nature Communications}, 14\penalty0 (1), 2023.

\bibitem[Seelig \& Jayaraman(2015)Seelig and Jayaraman]{neural_dynamics_landmark}
Seelig, J.~D. and Jayaraman, V.
\newblock Neural dynamics for landmark orientation and angular path integration.
\newblock \emph{Nature}, 521\penalty0 (7551), 2015.

\bibitem[Shen \& McNaughton(1996)Shen and McNaughton]{shen1996modeling}
Shen, B. and McNaughton, B.~L.
\newblock Modeling the spontaneous reactivation of experience-specific hippocampal cell assembles during sleep.
\newblock \emph{Hippocampus}, 6\penalty0 (6):\penalty0 685--692, 1996.

\bibitem[Skaggs \& McNaughton(1996)Skaggs and McNaughton]{replay_of_neuronal}
Skaggs, W.~E. and McNaughton, B.~L.
\newblock Replay of neuronal firing sequences in rat hippocampus during sleep following spatial experience.
\newblock \emph{Science}, 271\penalty0 (5257), 1996.

\bibitem[Song \& Ermon(2019)Song and Ermon]{generative_modeling_by_estimating_gradients}
Song, Y. and Ermon, S.
\newblock Generative modeling by estimating gradients of the data distribution.
\newblock In \emph{Advances in Neural Information Processing Systems (NeurIPS)}, volume~32, 2019.

\bibitem[Sorscher et~al.(2019)Sorscher, Mel, Ganguli, and Ocko]{sorscher2019unified}
Sorscher, B., Mel, G., Ganguli, S., and Ocko, S.
\newblock A unified theory for the origin of grid cells through the lens of pattern formation.
\newblock In Wallach, H.~M., Larochelle, H., Beygelzimer, A., d'Alch\'{e} Buc, F., Fox, E.~B., and Garnett, R. (eds.), \emph{Advances in Neural Information Processing Systems}, volume~32. Curran Associates, Inc., 2019.

\bibitem[Stachenfeld et~al.(2017)Stachenfeld, Botvinick, and Gershman]{hippocampus_predictive_map}
Stachenfeld, K.~L., Botvinick, M.~M., and Gershman, S.~J.
\newblock The hippocampus as a predictive map.
\newblock \emph{Nature neuroscience}, 20\penalty0 (11), 2017.

\bibitem[Stella et~al.(2019)Stella, Baracskay, O’Neill, and Csicsvari]{hippocampal_reactivation_random}
Stella, F., Baracskay, P., O’Neill, J., and Csicsvari, J.
\newblock Hippocampal reactivation of random trajectories resembling brownian diffusion.
\newblock \emph{Neuron}, 102\penalty0 (2), 2019.

\bibitem[Tang et~al.(2023)Tang, Barron, and Bogacz]{sequential_memory}
Tang, M., Barron, H., and Bogacz, R.
\newblock Sequential memory with temporal predictive coding.
\newblock In \emph{Advances in Neural Information Processing Systems (NeurIPS)}, volume~36, 2023.

\bibitem[Tang et~al.(2024)Tang, Barron, and Bogacz]{learning_grid_cells}
Tang, M., Barron, H., and Bogacz, R.
\newblock Learning grid cells by predictive coding.
\newblock \emph{arXiv preprint arXiv:2410.01022}, 2024.

\bibitem[Theodoni et~al.(2018)Theodoni, Rovira, Wang, and Roxin]{theodoni2018theta}
Theodoni, P., Rovira, B., Wang, Y., and Roxin, A.
\newblock Theta-modulation drives the emergence of connectivity patterns underlying replay in a network model of place cells.
\newblock \emph{eLife}, 7:\penalty0 e37388, 2018.
\newblock ISSN 2050-084X.

\bibitem[Tingley \& Peyrache(2020)Tingley and Peyrache]{tingley2020methods}
Tingley, D. and Peyrache, A.
\newblock On the methods for reactivation and replay analysis.
\newblock \emph{Philosophical Transactions of the Royal Society B: Biological Sciences}, 375\penalty0 (1799):\penalty0 20190231, 2020.

\bibitem[Tononi \& Cirelli(2014)Tononi and Cirelli]{sleep_price}
Tononi, G. and Cirelli, C.
\newblock Sleep and the price of plasticity: from synaptic and cellular homeostasis to memory consolidation and integration.
\newblock \emph{Neuron}, 81\penalty0 (1), 2014.

\bibitem[Uria et~al.(2022)Uria, Ibarz, Banino, Zambaldi, Kumaran, Hassabis, Barry, and Blundell]{Uria2020Spatial}
Uria, B., Ibarz, B., Banino, A., Zambaldi, V., Kumaran, D., Hassabis, D., Barry, C., and Blundell, C.
\newblock A model of egocentric to allocentric understanding in mammalian brains.
\newblock \emph{bioRxiv}, 2022.

\bibitem[Wood et~al.(2018)Wood, Bauza, Krupic, Burton, Delekate, Chan, and O’Keefe]{honeycomb_maze}
Wood, R.~A., Bauza, M., Krupic, J., Burton, S., Delekate, A., Chan, D., and O’Keefe, J.
\newblock The honeycomb maze provides a novel test to study hippocampal-dependent spatial navigation.
\newblock \emph{Nature}, 554\penalty0 (7690), 2018.

\bibitem[Xu et~al.(2025)Xu, Gao, Zhang, Wei, and Wu]{conformal_isometry}
Xu, D., Gao, R., Zhang, W., Wei, X.-X., and Wu, Y.~N.
\newblock On conformal isometry of grid cells: Learning distance-preserving position embedding.
\newblock In \emph{International Conference on Learning Representations (ICLR)}, 2025.

\bibitem[Xu et~al.(2012)Xu, Jiang, Poo, and Dan]{xu2012activity}
Xu, S., Jiang, W., Poo, M., and Dan, Y.
\newblock Activity recall in a visual cortical ensemble.
\newblock \emph{Nature Neuroscience}, 15\penalty0 (3):\penalty0 449–455, 2012.
\newblock ISSN 1546-1726.

\bibitem[Yang et~al.(2023)Yang, Zhang, Song, Hong, Xu, Zhao, Zhang, Cui, and Yang]{diffusion_models_comprehensive}
Yang, L., Zhang, Z., Song, Y., Hong, S., Xu, R., Zhao, Y., Zhang, W., Cui, B., and Yang, M.-H.
\newblock Diffusion models: A comprehensive survey of methods and applications.
\newblock \emph{ACM Computing Surveys}, 56\penalty0 (4), 2023.

\bibitem[Zhang et~al.(2023)Zhang, Zhang, Song, Yi, and Kweon]{survey_on_masked}
Zhang, C., Zhang, C., Song, J., Yi, J. S.~K., and Kweon, I.~S.
\newblock A survey on masked autoencoder for visual self-supervised learning.
\newblock In \emph{IJCAI}, 2023.

\bibitem[Zhang et~al.(2022)Zhang, Long, Zhang, and Chen]{excitatory_inhibitory_recurrent}
Zhang, X., Long, X., Zhang, S.-J., and Chen, Z.~S.
\newblock Excitatory-inhibitory recurrent dynamics produce robust visual grids and stable attractors.
\newblock \emph{Cell Reports}, 41\penalty0 (11), 2022.

\bibitem[Zhao et~al.(2023)Zhao, Yang, and Fung]{short-term_postsynaptic}
Zhao, H., Yang, S., and Fung, C. C.~A.
\newblock Short-term postsynaptic plasticity facilitates predictive tracking in continuous attractors.
\newblock \emph{Frontiers in Computational Neuroscience}, 17, 2023.

\bibitem[Ólafsdóttir et~al.(2015)Ólafsdóttir, Barry, Saleem, Hassabis, and Spiers]{olafsdottir2015hippocampal}
Ólafsdóttir, H.~F., Barry, C., Saleem, A.~B., Hassabis, D., and Spiers, H.~J.
\newblock Hippocampal place cells construct reward related sequences through unexplored space.
\newblock \emph{eLife}, 4:\penalty0 e06063, 2015.
\newblock ISSN 2050-084X.

\bibitem[Ólafsdóttir et~al.(2018)Ólafsdóttir, Bush, and Barry]{olafsdttir2018therole}
Ólafsdóttir, H.~F., Bush, D., and Barry, C.
\newblock The role of hippocampal replay in memory and planning.
\newblock \emph{Current Biology}, 28\penalty0 (1):\penalty0 R37–R50, 2018.
\newblock ISSN 0960-9822.

\end{thebibliography}

\newpage
\appendix
\onecolumn
\section*{Supplementary Material}
\section{Methods}
\label{appendix:methods}

\subsection{Experiments}
\label{appendix:methods_expts}

\subsubsection{Ornstein-Uhlenbeck}

Awake trajectories $s(t)$ are simulated with $\dt = 0.02, \sigma_s = 0.1, \sigma_0 = 0.2, \theta = 2, \mu = 5$ for $T = 100$ iterations.
We used \Cref{eq:score_ou_main} as the deterministic component of $\drt{}$ since a 1D Orstein-Uhlenbeck process admits a closed-form expression for $\frac{d}{dr(t)} \log p(r(t))$.
In \Cref{fig:1d_mean_std}, underdampening corresponds to $\friction=0.5$, while adaptation corresponds to $b_a = 1, \tau_a = 100$.
% https://colab.research.google.com/drive/1y2qpqG53kmDiLtgvAyzTWAReBu9p0wZd#scrollTo=Iy733GlK0-WS

\subsubsection{T-maze and triangle}

\paragraph{Task description.}
In the 2D T-maze and triangle tasks, we simulate directed random walks (Ornstein-Uhlenbeck processes) along several directions, and train the RNN to path-integrate these walks from their velocities.
In the T-maze task, there are two directions of travel: both start from the origin and go up, then one goes left and the other goes right (orange and purple, respectively, in \Cref{fig:paths}).
Meanwhile, in the triangle task there are six directions: denoting the three corners of the equilateral triangle as $A = (0,0), B = (1,0)$, and $C = (1/2, \sqrt{3}/2)$, respectively, these directions are $\overrightarrow{AB}, \overrightarrow{BC}, \overrightarrow{CA}, \overrightarrow{AC}, \overrightarrow{CB}, \overrightarrow{BA}$, shown in \Cref{fig:paths} as blue, red, gray, pink, green, and orange, respectively.
Awake and replay paths last $100$ timesteps, unless exploration is being measured, in which replay paths are allowed to last $400$ timesteps.

\paragraph{Architecture and training.}
For both tasks, we used shallow leaky-ReLU RNNs with linear output projections, as described in \Cref{appendix:discretization}.
In the T-maze task, we used 20 hidden neurons, while in the triangle task, we used 40.
We use masking difficulty $k=3$ in a progressive curriculum.
Triangle-task RNNs are trained with $k=1$ for $20,000$ epochs, then at $k=2$ and $k=3$ for $5,000$ epochs each; T-maze-task RNNs are trained likewise, but with fewer epochs ($12,000$ at $k=1$ and $5,000$ at $k=2$ and $k=3$ each).
In each task, we train 5 RNNs with different seeds.

\paragraph{Hidden state initialization.}
In each task, multiple directions start from the same point in space (for example, $\overrightarrow{AB}$ and $\overrightarrow{AC}$ in the triangle task), so we add onto initial hidden states some random vectors, orthogonal to the 2D output projection, specific to each direction.
For example, hidden states for $\overrightarrow{AB}$ and $\overrightarrow{AC}$ paths are both initialized to start near $A$, but $\overrightarrow{AB}$ paths also start with a fixed random vector $\bm{\eta}_{AB}$ added to their initial hidden states, whereas $\overrightarrow{AC}$ have a different fixed random vector $\bm{\eta}_{AC}$ added to their initial hidden states.
This notion of initializing hidden states in directions orthogonal to output projections has been previously discussed in computational neuroscientific contexts \citep{preparatory_activity}, and recent evidence suggests that memories (which replay is similar to) have uniquely identifiable, output-orthogonal patterns of neural activation \citep{barcoding_episodic}.

\subsubsection{Rat place cell trajectories}

Our unbiased (undirected random walks within a 2D box) and biased (directed random walks that head towards the center of a 2D box) rat place cell trajectory experiments are identical to those of \cite{sufficient_conditions}, with 5 different ReLU RNN seeds per task, but with two modifications:
\begin{enumerate}
\item We add masked training, with difficulties $k=3$ in the biased task and $k=6$ in the unbiased task.
We use a higher masking difficulty in the unbiased task to encourage longer replay trajectories; at lower values of $k$, unbiased replay paths are much shorter than awake trajectories.
In Langevin sampling terms, we conjecture this is because the score function $\score$ is fairly weak since unbiased trajectories are uniformly distributed\footnote
{
This relative weakness the unbiased activity score function (especially compared to the biased activity score function, where trajectories all go towards the center of a 2D box) is also why we use relatively small adaptation strengths $b_a$ in simulating replay from unbiased-task RNNs.
}.
\item We use a slower, but more detailed, decoder. \cite{sufficient_conditions} decode position at a given timestep as the average position of the top 3 most active place cells, which is fast but effectively quantizes decoded positions.
We use this method to initialize our positions, but then we optimize our positions via gradient descent to minimize the mean-squared error between observed place cell activity and the place cell activity that would correspond to these optimized positions (this is made possible by our knowledge of the exact place cell activity function). 
Our procedure takes much longer, but results in truly continuous replay trajectories, avoiding any significant quantization.
\end{enumerate}

\subsection{Hyperparameters}
\label{appendix:hyperparameters}

\subsubsection{Activation function}

In our work, we have mostly used RNNs with ReLU or leaky ReLU activation functions (except in \Cref{appendix:tanh}); our reasons for doing so are threefold.
The first is that ReLU or ReLU-like activations, unlike tanh activations, do not always saturate. This can mitigate vanishing gradients in RNNs (although care must be taken to avoid exploding gradients), but more importantly we have observed that this reduces the strength of fixed attractors learned by path-integrating RNNs, which is essential for exploration---in \Cref{appendix:tanh} we observe that replay in saturating RNNs explores less.
The second is that, compared to RNNs with multiplicative gating interactions (e.g., GRUs and LSTMs), ReLU RNNs are more biologically plausible insofar as they can be easily interpreted as the nonnegative, sparse firing rates of spiking networks.
The third is that most other previous works in sequential predictive coding tend to use ReLU activations, and have shown no significant differences in results when using more complex nonlinearities:
\begin{itemize}
\item \citet{predictive_coding_consequence}, some of the first authors to report the emergence of predictive representations emerging in RNNs, use ReLU activations.
\item \citet{sufficient_conditions} primarily use ReLU networks, and report similar results with GRU networks.
\item \citet{sequential_predictive_learning} use ReLU activations in conjunction with layer normalization, but report similar results when layer normalization is removed. 
\item \citet{predictive_sequence_learning} use tanh for must results, but when they care about the emergence of sparse, localized, nonnegative activations (i.e., place cells), they use ReLU activations.
\item \citet{sorscher2019unified, conformal_isometry}, and \citet{excitatory_inhibitory_recurrent} were all interested in the emergence of grid cells in predictive representations and primarily used ReLU or ReLU-like activations; those that tried using other nonlinearities did not observe notable differences in results.
\item \citet{learning_grid_cells}, who were also interested in grid cells, use both ReLU and tanh activations, and also report no notable differences between the two.
\item \citet{sequential_memory} use linear networks, and notice no significant difference when using tanh activations.
\end{itemize}

\subsection{Metrics}

\subsubsection{Wasserstein distance}
We seek to compare distributions of trajectories. 
In most of our tasks, these are objects of dimension $T \times 2$, where $T$ is the number of timesteps (recall that T-maze and triangle tasks are already in 2D, while rat trajectories are analyzed after being projected into 2D space).
To compare such high-dimensional distributions, we use Wasserstein distances \citep{statistical_aspects_wasserstein}:
\begin{itemize}
\item In the T-maze and triangle tasks, we first calculate the Wasserstein distance between awake and replay paths belonging to the same direction (e.g., $\overrightarrow{AB}$ in the triangle task).
Since paths along the same direction should resemble, if not obey, Gaussian processes, these paths should approximately be normally distributed, and so we can use the closed-form equation for the Wasserstein distance between two normal distributions.
After computing Wasserstein distances for each direction, we take the average as our final distance.
KL divergence might have also worked in this task, but in practice the covariance matrices of the $(T \times 2)$-dimensional distributions were singular.
\item In the rat experiments, decoded 2D trajectories are not easily decomposed into groups of Gaussian processes, so instead we apply sliced Wasserstein distances \citep{sliced_radon_wasserstein} to compare the $(T \times 2)$-dimensional distributions.
Sliced Wasserstein distances are essentially calculated by taking many random projections of two distributions onto 1D, where Wasserstein distances have a closed form.
Computing KL divergences instead would have been challenging, since rat path distributions are high-dimensional and do not admit straightforward estimations of probability density.
\end{itemize}

\subsubsection{Reach times}

In the T-maze, triangle, and biased rat tasks, awake trajectories have clearly defined endpoints that they reach.
Replay trajectories also aim for these endpoints, although second-order dynamics modifiers like adaptation and underdampening affect how quickly or how closely replay paths reach them.
We quantify these changes by measuring the timesteps required for replay paths to get within 10\% of their endpoints (for example, the timesteps required for $\overrightarrow{AB}$ paths to get within a 0.1$|\overrightarrow{AB}|$ of $B$).

\subsubsection{Path lengths}

One way we quantify exploration in replay paths is through path length.
We calculate this simply as the sum of velocity magnitudes.

\subsubsection{Regions visited}

Another way we quantify exploration in the T-maze and traingle tasks is through \textit{regions visited}.
As explained in \Cref{fig:region_count_avg}, \textit{regions} are portions of input space where all points are closest to the same endpoint.
In the triangle task, there are 3 endpoints (which all also act as initial points), while in the T-maze there are 2 endpoints and 1 shared starting point, which for the purposes of region assignment we also consider an ``endpoint''.
Thus, in each task there are 3 endpoints, and thus 3 regions.
Awake trajectories go from one region to another and stay there, but replay trajectories might proceed to visit more regions.
We define a ``visit'' as a contiguous presence within a single region for at least 10 timesteps.

\subsubsection{Additional exploration metrics}

\begin{figure}
\centering
\includegraphics[width=0.9\linewidth]{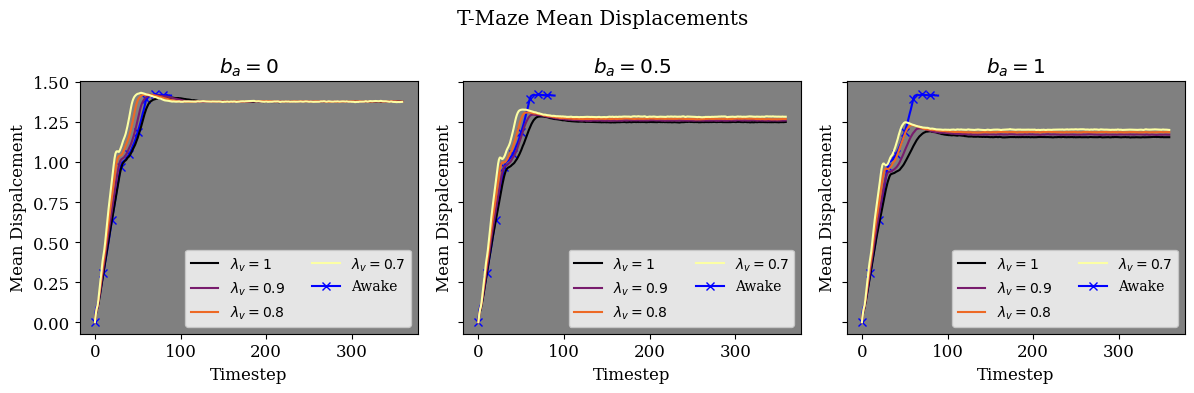}
\includegraphics[width=0.9\linewidth]{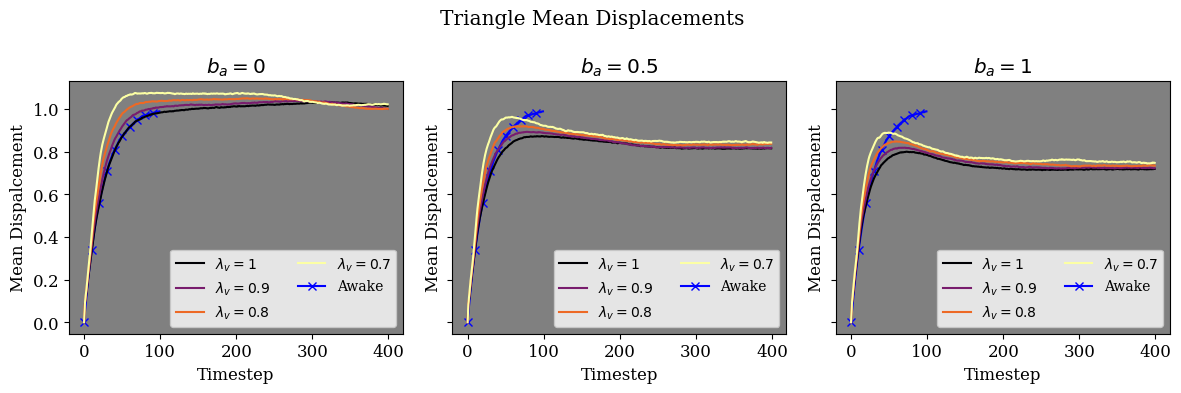}
\includegraphics[width=0.9\linewidth]{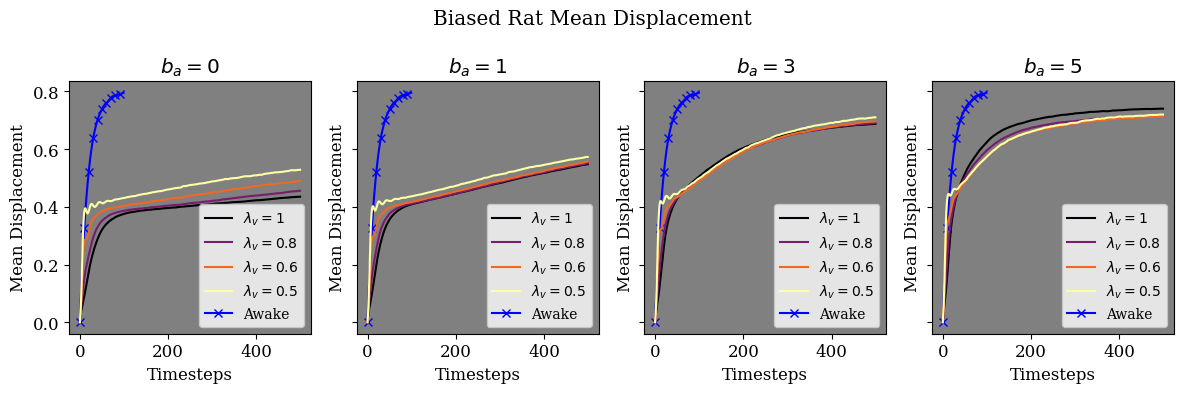}
\includegraphics[width=0.9\linewidth]{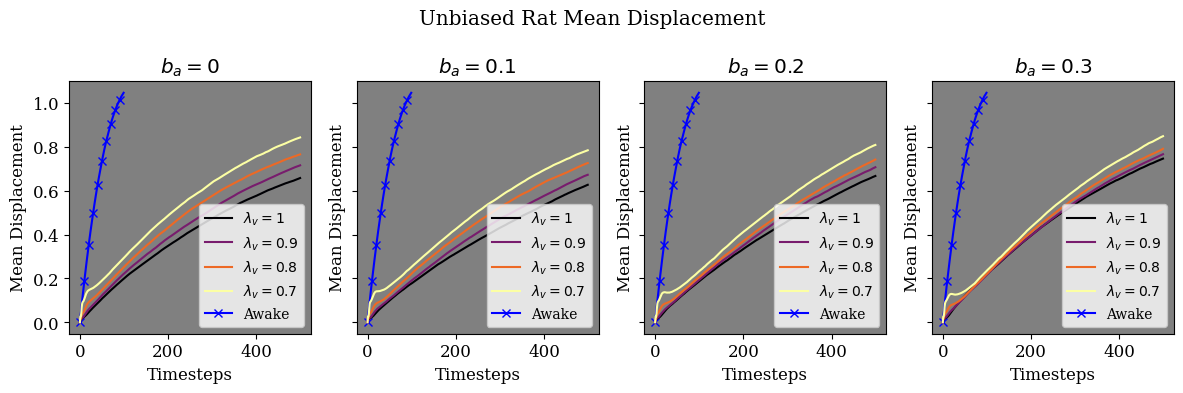}
%\includesvg[width=0.9\linewidth]{img/mean_displacement/tmaze_md.svg}
%\includesvg[width=0.9\linewidth]{img/mean_displacement/triangle_md.svg}
%\includesvg[width=0.9\linewidth]{img/mean_displacement/biased_md.svg}
%\includesvg[width=0.9\linewidth]{img/mean_displacement/unbiased_md.svg}
\caption{
\textbf{Underdampening ($\friction < 1$) increases mean displacement of replay trajectories, especially at early timesteps.}
Similarly to \cite{flexible_modulation}, here we analyze the \textit{mean displacement}, or distance, of replay trajectories from their starting points as a function of time.
For reference, mean displacement over time is also plotted for awake trajectories.
In exploration experiments, we simulate replay for $4\times$ the duration of awake trajectories.
Higher mean displacements over time generally correlate with increased path length and exploration.
Note that awake activity is always calculated with $b_a=0, \friction=1$, hence why awake statistics are the same across plots within the same task.
}
\label{fig:mean_displacement}
\end{figure}

\begin{figure}
\centering
\includegraphics[width=0.9\linewidth]{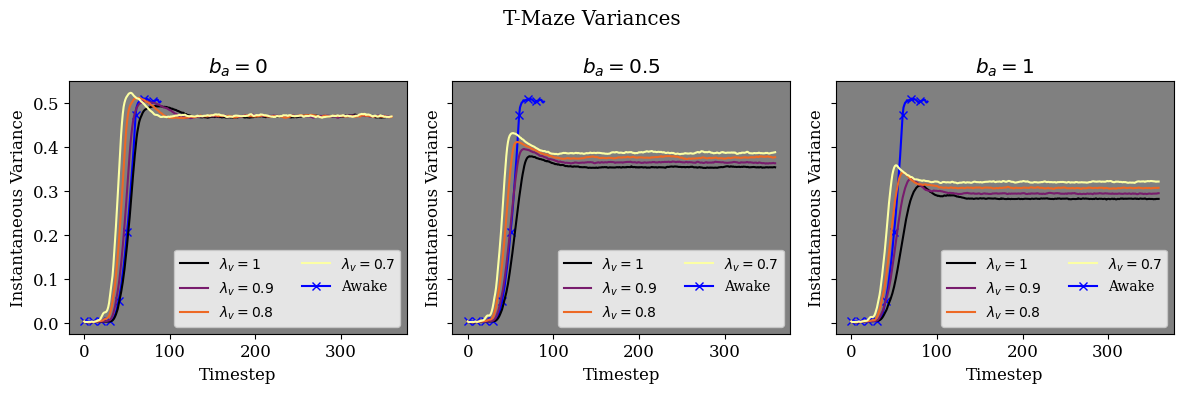}
\includegraphics[width=0.9\linewidth]{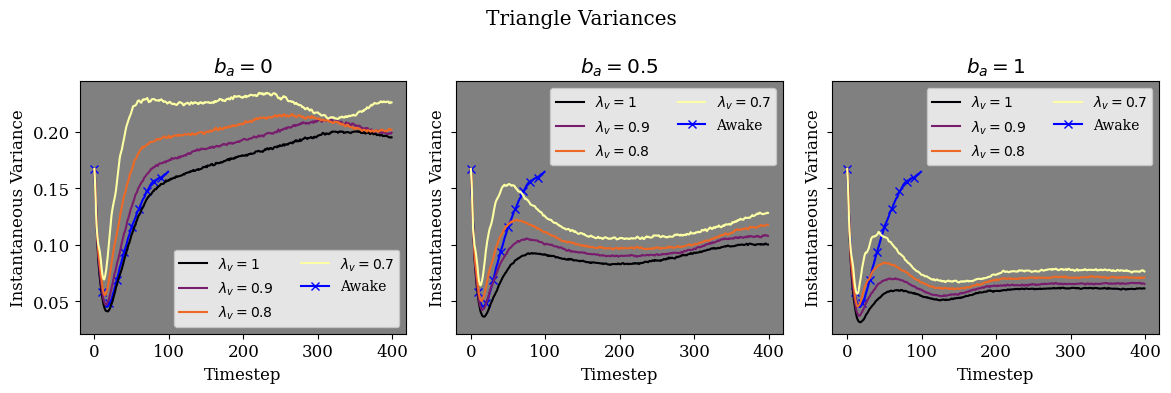}
\includegraphics[width=0.9\linewidth]{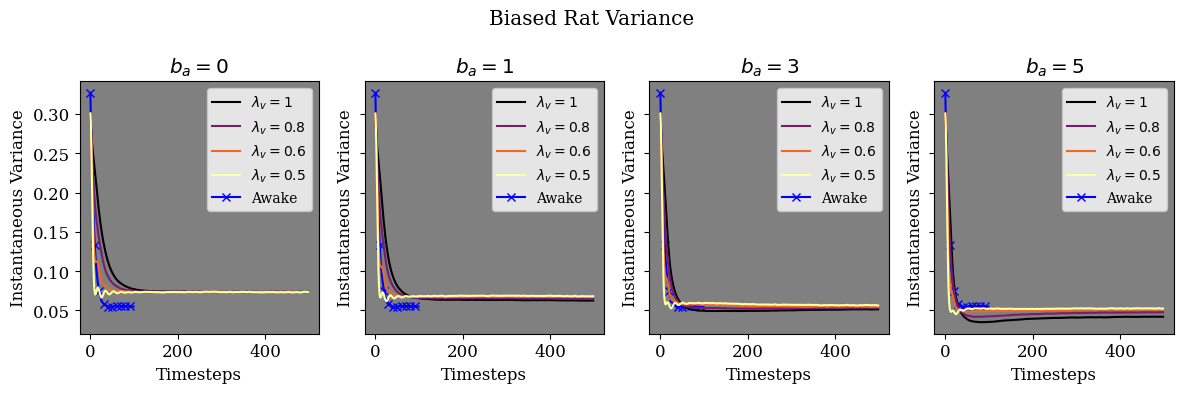}
\includegraphics[width=0.9\linewidth]{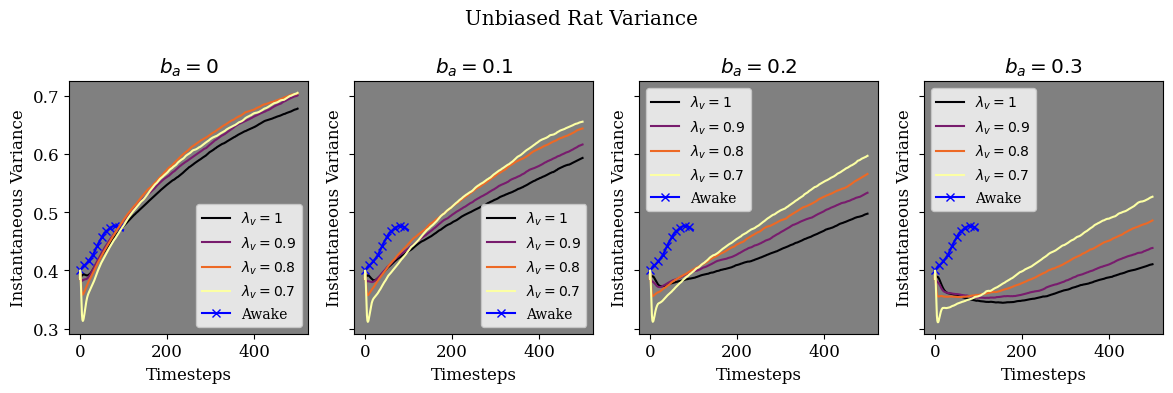}
%\includesvg[width=0.9\linewidth]{img/time_variances/tmaze_var.svg}
%\includesvg[width=0.9\linewidth]{img/time_variances/triangle_var.svg}
%\includesvg[width=0.9\linewidth]{img/time_variances/biased_var.svg}
%\includesvg[width=0.9\linewidth]{img/time_variances/unbiased_var.svg}
\caption{
\textbf{Underdampening ($\friction < 1$) increases replay trajectory variances.}
Like in \Cref{fig:mean_displacement}, here we analyze the variance of trajectories at each timestep.
Trajectories are 2D, so the variances plotted above are simply the averages of the variances along each coordinate.
Underdampening increasing variance is another confirmation that underdampening complements adaptation-induced exploration.
}
\label{fig:time_variances}
\end{figure}

For completeness, we add two other measures of exploration: \emph{mean displacement}, proposed by \citet{flexible_modulation}, and variance.
They are simply defined as $\mathbb{E}_{p(\brt)} [ |\brt - \br(0)| ]$ and $\mathbb{E}_{p(\brt)} [\mathrm{Var}(\brt)]$.
Higher mean displacements and variances generally correlate with increased path length and exploration.
We plot them in \Cref{fig:mean_displacement} and \Cref{fig:time_variances}.

\subsection{Discretization and Implementation of Noisy RNNs}
\label{appendix:discretization}

While biological neural networks are continuous-time systems, RNNs are in practice implemented discretely.
In order to incorporate second-order processes like adaptation and underdamped sampling in RNNs, we must discretize them, which we do as follows:
\begin{align}
    \Delta \widetilde{\br}(t) &= f(\brt, \but, \sigmaeta) - \brt
    \\
    \bm{c}(t + \dt) &= \bct + \frac{1}{\tau_a} (-\bct + b_a \brt)
    \\
    \bm{v}(t + \dt) &= (1 - \friction) \bvt + \Delta \widetilde{\br}(t)
    \\
    \br(t + \dt) &= \brt -\bct + \bm{v}(t + \dt)
\end{align}

We train and sample from ReLU RNNs: $f(\brt, \but, \sigmaeta) = \kappa \brt +  \mathrm{ReLU}(\bm{W}_r \brt + \bm{W}_{in} \but + \sigmaeta)$ (where $k \in [0, 1]$ is also learnable) unless otherwise stated (a choice justified in \Cref{appendix:hyperparameters}).
In the absence of directed inputs $\but$, $f(\brt, \but, \sigmaeta)$ should be roughly equivalent to the noisy score function from Equation \ref{eq:rnn_langevin}.

If the adaptation strength $b_a = 0$, then there is no adaptation ($\bct = \bm{0}$).
As for our \textit{friction} term $\friction \in [0, 1]$, if $\friction = 1$, then there is no underdamped sampling since $\bvt = \drt{} + \sigmaeta$ no longer accumulates previous values of $\bvt$.
In other words, the friction term $\friction$ allows for smooth interpolation between overdamped and underdamped sampling.

Note that, unlike $\bm{c}(t + \dt)$ and $\bm{v}(t + \dt)$, $\br(t + \dt)$ depends on another term calculated at $t + \dt$. This is a symplectic Euler discretization of the second-order dynamics, which we employ for to ensure the stability of interactions between $\brt$ and its momentum $\bvt$.
The negative feedback $\bct$ is much more stable, so its discretization goes unchanged.

Across all trained RNN experiments we use $\tau_a = 0, \friction=1$ for awake activity (i.e., training).
When generating replay, we always use $\tau_a = 100$, and either $T=100, 400,$ or $500$ timesteps, depending on whether we are measuring fidelity and speed ($T=100$, like in Figures \ref{fig:paths}, \ref{fig:wd}, \ref{fig:reachtime_heatmaps}) or exploration ($T=400$ for T-maze and triangle, $T=500$ for rat tasks, like in Figures \ref{fig:exploration_metrics}, \ref{fig:region_count_avg}).

\subsection{Deviations from Traditional Langevin Sampling}
\label{appendix:deviations_langevin}

We must note a few minor distinctions between our replay RNNs and traditional Langevin sampling.

\paragraph{Stationarity.}
We treat neural replay as a sequence of events, and are thus interested in the joint distribution of replay activity at all timesteps $p\left(\{\brt\}_{t=1}^T\right)$.
This distribution, however, is not necessarily stationary across $t$. Stationarity would imply that $p(\brt)$ has no need or intention of traversing along any meaningful path.
While a path-integrating RNN does perform gradient ascent along $\log p(\brt)$, it does so in a piecewise manner along $t$ rather than jointly along all $t$ simultaneously.
This variation on gradient ascent, in combination with the non-stationarity of $p(\brt)$, means that Langevin sampling guarantees do not hold.
For this reason, we make modifications to our RNNs that under stationary Langevin sampling theory might seem unprincipled.

\paragraph{Underdampening.}
The friction term $\friction$ applied to the velocity $\bvt$ is subtly different from $\gamma$ in Equations \ref{eq:underdamped_langevin_1st_order} and \ref{eq:underdamped_langevin_2nd_order}.
$\friction \in [0, 1]$ can be interpreted as trying to map $\gamma \in [0, \infty)$ onto $[0, 1]$.
Like $\gamma \rightarrow \infty$, $\friction \rightarrow 1$ removes any dependency of $\bm{v}(t + \dt)$ on $\bvt$.

\paragraph{Noise scaling.}
RNN replay, unlike Langevin sampling, is sensitive to the variance of noise used.
We found that omitting the $\sqrt{2}$ factor in front of $\sigmaeta$ worked best.
For this reason, we also omitted $\friction$ as a scaling term on $\sigmaeta$.
If $\friction$ is allowed to modulate $\sqrt{2} \sigmaeta$, we found that it is easy to show improvements in replay fidelity when $\friction < 1$, but we found these to come from noise scaling rather than from underdampening or momentum as mechanisms.

\paragraph{Relation to diffusion models.}
Throughout this work we treat RNNs as generative models using a variant of Langevin sampling.
Expressive generative models, in particular \textit{diffusion models} whose activity resembles Langevin sampling, have received much attention from the machine learning community \citep{self_consuming, boomerang, diffusion_models_comprehensive, diffusion_models_in_vision, diffusion-lm, chatgpt_is_not, generative_modeling_by_estimating_gradients}.
The fundamental differences between our RNNs and diffusion models are twofold.
The first is that each timestep of a generated replay trajectory is generated sequentially and with only one effective step along the gradient of log-likelihood, whereas a diffusion model would generate all timesteps of a path simultaneously and with many steps along the gradient of log-likelihood.
The second is that diffusion models use time-varying noise levels (annealed dynamics) and are explicitly conditioned on time as an input, whereas our RNNs use the same noise level across time and are never explicitly conditioned on time.
\section{Overview of Background}
\label{appendix:background_overview}

Throughout the text we make use of notions including sampling, Langevin dynamics, replay, path-integration, awake and quiescent activity, adaptation, and masked training.
Here we gather these notions, which have been defined in various previous work, as follows:
\begin{enumerate}
\item Our work examines how trained noisy RNNs can act as generative models in the absence of inputs, which prior work including \citet{sufficient_conditions} consider to be models of replay in neuronal circuits.
Accordingly, we define Langevin dynamics, with (\Cref{eq:overdamped_langevin}) and without (Equations \ref{eq:underdamped_langevin_2nd_order} and \ref{eq:underdamped_langevin_1st_order}) momentum, as an iterative process by which an agent could attempt to sample from an unknown distribution $p(x)$ using only knowledge of its \textit{score function} $\nabla \log p(x)$, which can be learned from noisy observations drawn from the distribution. The following points provide more detail on the sampling process and why Langevin dynamics arise:
\begin{itemize}
    \item Biological networks such as those underlying navigation must accurately estimate environmental state variables such as position from observations of self-motion, even in the presence of intrinsic noise.
    \item This noise induces a distribution over network states, i.e., activity, however, it is unknown exactly what the true noise distribution is.
    \item Accurate state estimation would thus require the network to optimally remove intrinsic noise without access to the exact parameters of the noise distribution, but given noisy network states.
    \item This optimal solution can be accomplished by learning the score function of noisy network states and using it to denoise network states over time as additional inputs and noise are presented \citep{miyasawa1961empirical}.
    \item When the network receives no inputs, its noise-driven dynamics still attempt to remove intrinsic noise, but the learned score function is for the distribution of waking task-like activity. This causes the network states to resemble actual samples drawn from the distribution over network states in the presence of inputs. The theoretical results show that the network dynamics during this quiescent state carry out sampling by Langevin dynamics, i.e., using the score function of the network's activity distribution during task performance to yield quiescent network activity that resembles waking activity.
\end{itemize}
\item Biological neural circuits like the hippocampus can navigate through environments when awake, and recall navigatory episodes during rest. Accordingly, we summarize the proof from \citet{sufficient_conditions} that a noisy RNN trained on a biologically relevant navigation task like path-integration can act as a generative model. 
\begin{itemize}
\item Definitions 2.1 and 2.2 simply express that our RNNs have nonlinear dynamics and internal noise, and are trained to estimate a signal $\bst$ by integrating observations of $\boldsymbol{s}'(t)$. This phenomenon, known as path-integration, underlies navigation and is used to estimate one's position from self-motion cues.
\item Assumption \ref{assumption:additive_decomposition} asserts that RNN dynamics can be decomposed into additive components. 
Assumption \ref{assumption:gaussian_r(t)} simply states that the optimal RNN dynamics would lead to accurate path-integration, such that the conditional probability distribution over network states (activity) given environmental states (position) would be Gaussian and identical to the additive noise in our networks. That is, the variability over network states would only be due to intrinsic noise in the system and have the same statistics. 
Finally, Assumption \ref{assumption:greedy} assumes that the RNN dynamics are greedily optimal for path-integration. Greedy optimization is a sensible way of partitioning effort across time in path-integration: the network does the best that it can at each timestep, assuming that at each previous timestep the best possible job has been done.
\item Theorem \ref{theorem:optimal_update} states that path-integrative RNN dynamics will use the score function of waking activity to remove intrinsic noise from the system and use inputs to update state estimates. This is a two-step greedily optimal solution for path integration in the presence of intrinsic noise: first, intrinsic noise must be removed, following which the inputs must be used to update the estimate of the current environmental state. 
Theorem \ref{theorem:langevin_replay} shows that these dynamics will result in Langevin sampling from the distribution of waking activity in the absence of any inputs, i.e., analogous to sleep. This means that the quiescent network dynamics sample neural states from the same distribution as waking, task-like activity, and can thus represent sequences like those during awake task performance, i.e., leading to replay. 
\end{itemize}
\item Finally, we define existing methods of modulating RNN activity or training that affect RNN replay distributions: adaptation (negative feedback), which encourages diversity in replay paths, and masked training, which encourages coherence in replay.
\end{enumerate}
\section{Score Functions of Gaussian Distributions}
\label{sec:appendix_score}

For any matrix calculus involved, we use denominator layout.

\subsection{Multivariate Gaussian Distribution}

Let's suppose $\br \sim \mathcal{N}(\bmu, \bSigma)$.
If $\br \in \mathbb{R}^d$, then:
\begin{align}
    p(\br) &= \frac{1}{\sqrt{(2 \pi)^d |\bSigma|}} \exp \left( -\frac{1}{2} (\br - \bmu)^T \bSigma^{-1} (\br - \bmu) \right)
    \\
    \log p(\br) &\propto -\frac{1}{2} (\br - \bmu)^T \bSigma^{-1} (\br - \bmu)
    \\
    \nabla_{\br} \log p(\br) &= -\frac{1}{2} ((\bSigma^{-1})^T + \bSigma^{-1}) (\br - \bmu)
    \\
    &= -\bSigma^{-1} (\br - \bmu)
    \label{eq:gaussian_score}
    \\
    &= -\sigma^{-2}(r - \mu) \enspace \mathrm{if} \enspace r \in \mathbb{R}
\end{align}

\subsection{Score Function of \texorpdfstring{$r(t)$}{r(t)} for Gaussian \texorpdfstring{$s(t)$}{s(t)}}
\label{sec:appendix_score_of_r(t)}

Recall that $p(\brt | \bst) \sim \mathcal{N}(\bDdag \bst, \, \bI \sigmadt)$ from Equation \ref{eq:p_r|s}.
If we suppose that $\bst$ is normally distributed with mean and covariance $\bmu_{\bst}, \bSigma_{\bst}$, then we can obtain $p(\brt)$:
\begin{equation}
    p(\brt) \sim \mathcal{N}(\bDdag \bmu_{\bst}, \, \bI \sigmadt + \bDdag \bSigma_{\bst} (\bDdag)^T),
\end{equation}
which we can plug into Equation \ref{eq:gaussian_score} to get $\score$:
\begin{equation}
    \score =
    -\left(
    \bI \sigmadt +
    \bDdag \bSigma_{\bst} (\bDdag)^T
    \right)^{-1}
    \left(
    \brt - \bDdag \bmu_{\bst}
    \right)
\end{equation}
Moreover, we can use the above score function to calculate the optimal $\Delta \br(t + \dt)$ in Equation \ref{eq:delta_r_1_and_2_star}:
\begin{equation}
\begin{split}
    \Delta \br^*(t + \dt) &=
    \sigmadt \left(
    \bI \sigmadt +
    \bDdag \bSigma_{\bst} (\bDdag)^T
    \right)^{-1}
    \left(
    -\brt + \bDdag \bmu_{\bst}
    \right)
    \\
    &+ \bDdag \bs'(t) \dt + \sigmaeta
\end{split}
\end{equation}
Some properties of the leakage matrix $\sigmadt \left( \bI \sigmadt + \bDdag \bSigma_{\bst} (\bDdag)^T \right)$ include:
\begin{enumerate}
    \item The covariance matrix $\bSigma_{\bst}$ is positive semidefinite (PSD): all its eigenvalues are $\geq 0$.
    \item $\bDdag \bSigma_{\bst} (\bDdag)^T$ is also PSD
    \footnote{
    Proof: If $\bm{B}$ is PSD, then $x^T \bm{ABA}^T x = (\bm{A}^Tx)^T \bm{B} (\bm{A}^T x) = v^T \bm{B} v \geq 0$.
    }, symmetric, and therefore diagonalizable.
    \item The eigenvalues of $(\bI \sigmadt + \bDdag \bSigma_{\bst} (\bDdag)^T)^{-1}$ are thus all $\leq (\sigmadt)^{-1}$
    \footnote{
    Proof: For diagonalizable $\bm{A}$, the $i$-th eigenvalue of $(\lambda \bI + \bm{A})^{-1} = (\bm{Q}(\lambda \bI + \bm{\Lambda})\bm{Q}^{-1})^{-1}$ is equal to $(\lambda + \bm{\Lambda}_{ii})^{-1}$, which can be no larger than $\lambda^{-1}$ if $\bm{A}$ is PSD.
    }. 
    \item The eigenvalues of $ \sigmadt (\bI \sigmadt + \bDdag \bSigma_{\bst} (\bDdag)^T)^{-1}$ are thus all $\leq 1$.
    \item If the off-diagonal entries of the leakage matrix above are sufficiently small in magnitude, then all the diagonal entries should be less than 1, as justified by the Gershgorin Circle Theorem.
    In fact, if the leakage matrix is diagonal, then it must have all values less than 1 (which could be achieved via sigmoid functions or perhaps spectral normalization).
    \item As for interpretation, smaller leakage eigenvalues means higher eigenvalues of $\bDdag \bSigma_{\bst} (\bDdag)^T$, or essentially, more noise.
    The maximum determinant of the leakage matrix is 1, when there is essentially no noise in $\bs(t)$.
\end{enumerate}
\section{Additional Score Function Results}
\label{appendix:more_score}

\paragraph{Wiener Processes.}
One simple stochastic process is the Wiener process, which in terms of navigation represents an undirected random walk ($\theta = 0$).
Assuming $s_w(0) = 0$, then $s_w(t) \sim \mathcal{N}(0, \sigma_s^2 t)$, and therefore $p(r_w(t)) \sim \mathcal{N}(0, \sigma_s^2 t + \sigmadt)$ from Equation \ref{eq:p_r|s}, producing the following score:
\begin{equation}
\sigmadt \, \nabla_{r_w(t)} \log p(r_w(t)) = \sigmadt \frac{-r_w(t)}{\sigma_s^2 t + \sigmadt} 
\end{equation}
Even from a simple Wiener process, we observe that the per-timestep optimal score is not constant with respect to $t$: at $t=0$, it equals $-r_{w}(t)$, while as $t$ approaches $\infty$, it approaches $0$.

\paragraph{Ornstein-Uhlenbeck Processes.}

Now we incorporate non-zero leakage ($\theta > 0$) to describe a directed random walk navigating from an arbitrary starting point $s_{ou}(0)$ towards a mean destination $\mu$.
If $p(s_{ou}(0)) \sim \mathcal{N}(0, \sigma_0^2)$, then $p(s_{ou}(t)) \sim \mathcal{N} \left(\mu (1 - e^{-\theta t}), \frac{\sigma_s^2}{2 \theta} (1 - e^{-2 \theta t}) + \sigma_0^2 e^{-\theta t} \right)$, and the score function is:
\begin{equation} \label{eq:score_ou_appendix}
\sigmadt \, \nabla_{r_{ou}(t)} \log p(r_{ou}(t)) = \sigmadt \frac{
-(r_{ou}(t) - \mu(1 - e^{-\theta t}))
}{
\frac{\sigma_s^2}{2\theta}(1 - e^{-2\theta t}) + \sigma_0^2 e^{-\theta t} + \sigmadt
}
\end{equation}
\section{Adaptation as a Second-Order Stochastic Differential Equation}
\label{sec:appendix_adaptation_sde}

Let us first combine the following two coupled linear stochastic differential equations into one second-order equation: 
\begin{align}
    dX_t &= (AX_t + BY_t + M)dt + \sigma dB_t, \quad dY_t = (CX_t + DY_t)dt
    \\
    d^2 X_t &= AdX_t + BdY_t + \sigma d(dB_t)
    \\
    &= AdX_t + BdY_t + \sigma d^2 B_t
    \\
    &= AdX_t + B(CX_t + DY_t)dt + \sigma d^2 B_t, \enspace
    Y_t = \frac{1}{Bdt} ( dX_t - \sigma dB_t - AX_t dt - M dt)
    \\
    &= AdX_t + BCX_t dt + BD \frac{1}{B dt} (dX_t - \sigma dB_t - AX_t dt - M dt)dt + \sigma d^2 B_t
    \\
    &= AdX_t + BCX_t dt + D(dX_t - \sigma dB_t - AX_t dt - M dt) + \sigma d^2 B_t
    \\
    &= (A+D) dX_t + (BC - AD)X_t dt - DM dt - \sigma D d B_t + \sigma d^2 B_t
\end{align}
Replacing all variables involved (except $dt, \sigma$) with matrices and vectors yields the same equation as long as $\bm{B}$ is invertible:
\begin{equation}
    d^2 \bm{x}_t = (\bm{A} + \bD) d\bm{x}_t + (\bm{BC} - \bm{AD}) \bm{x}_t dt - \bm{Dm} dt - \sigma \bD d\bm{B}_t + \sigma d^2 \bm{B}_t
\end{equation}
For consistency with the notation used throughout the paper, the equation above can be written as:
\begin{equation}
    \bm{x}''(t) = (\bm{A} + \bD) \bm{x}'(t) + (\bm{BC} - \bm{AD}) \bm{x}(t) - \bm{Dm} - \sigma \bD \bm{\eta}(t) + \sigma \bm{\eta}'(t)
\end{equation}
If we apply the following substitutions from Equations \ref{eq:gaussian_score} and \ref{eq:rnn_langevin_adaptation}:
\begin{itemize}
    \item $\bm{x}(t) \leftarrow \brt$,
    \item $\bm{A} \leftarrow -\sigmadt \bSigma^{-1}$,
    \item $\bm{B} \leftarrow -\bI$,
    \item $\bm{m} \leftarrow \sigmadt \bSigma_t^{-1} \bmu$,
    \item $\bm{C} \leftarrow -\frac{1}{\tau_a} \bI$,
    \item $\bD \leftarrow \frac{b_a}{\tau_a} \bI$,
\end{itemize}
then $\br''(t)$ is:
\begin{equation}
\begin{split}
    \br''(t)
    = 
    \left(
    \frac{b_a}{\tau_a} \bI - \sigmadt \bSigma^{-1}
    \right) \br'(t)
    \\
    +
    \left(
    \frac{1}{\tau_a} \bI + \frac{b_a}{\tau_a} \sigmadt \bSigma^{-1}
    \right) \brt
    \\
    -
    \frac{b_a}{\tau_a} \sigmadt \bSigma^{-1} \bmu
    -
    \sigma \frac{b_a}{\tau_a} \bm{\eta}(t) + \sigma \bm{\eta}'(t)
\end{split}
\end{equation}
Recall that, if $\brt$ follows a stationary Gaussian distribution, then $\score = \bSigma^{-1} (-\brt + \bmu)$ (Equation \ref{eq:gaussian_score}), and therefore $\frac{d^2}{d\brt^2} \log p(\brt) = - \bSigma^{-1}$.
Then,
\begin{equation}
\begin{split}
    \br''(t)
    =
    \left(
    \frac{b_a}{\tau_a} \bI + \sigmadt \frac{d^2}{d\brt^2} \log p(\brt)
    \right) \br'(t)
    \\
    -
    \frac{b_a}{\tau_a} \sigmadt~ \score + \frac{1}{\tau_a} \brt
    \\
    -
    \sigma \frac{b_a}{\tau_a} \bm{\eta}(t) + \sigma \bm{\eta}'(t)
\end{split}
\end{equation}
\section{Additional Results}
\label{appendix:more_results}

Here we present additional results or figures that supplement those of the main text.

\input{tikzfigs/paths_rat}

%\newcommand{\rtMinTmaze}{-30}
%\newcommand{\rtMaxTmazeHalf}{60}
%\newcommand{\rtMaxTmaze}{120}
%\newcommand{\rtMinTriangle}{-30}
%\newcommand{\rtMaxTriangleHalf}{55}
%\newcommand{\rtMaxTriangle}{110}
%\newcommand{\rtMinBiased}{-60}
%\newcommand{\rtMaxBiasedHalf}{150}
%\newcommand{\rtMaxBiased}{300}
%\pgfplotsset{
%    colormap={bwrTmaze}{
%        color(\rtMinTmaze)=(blue)
%        color(0)=(white)
%        color(\rtMaxTmaze)=(red)},
%    colormap={bwrTriangle}{
%        color(\rtMinTriangle)=(blue)
%        color(0)=(white)
%        color(\rtMaxTriangle)=(red)},
%    colormap={bwrBiased}{
%        color(\rtMinBiased)=(blue)
%        color(0)=(white)
%        color(\rtMaxBiased)=(red)}
%}

%% -----------------------
%% all reach time heatmaps
%% -----------------------
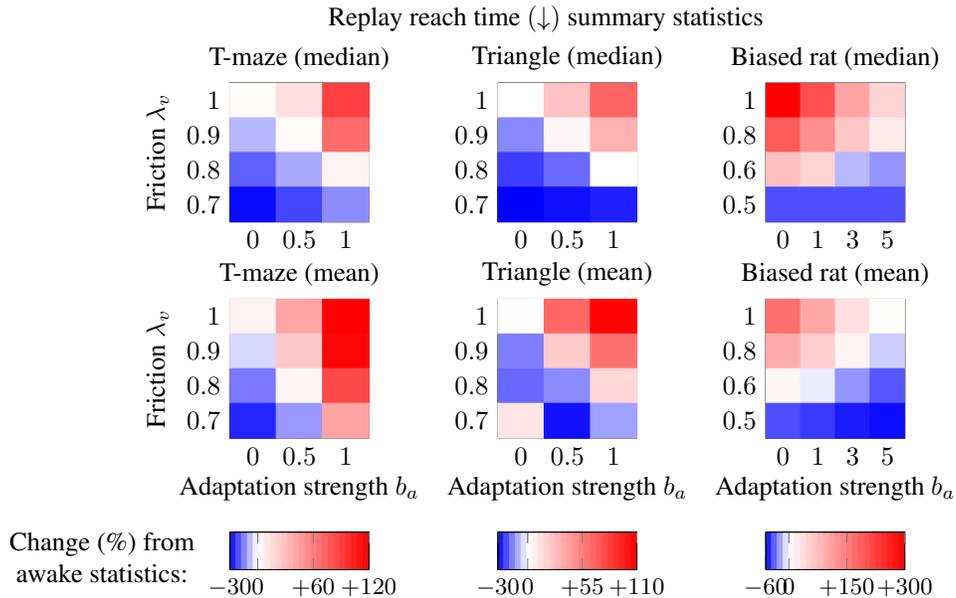
\begin{figure}[h]
\centering
\begin{tikzpicture}

\begin{groupplot}[
    group style = {
        group size = 3 by 2,
        horizontal sep = 0.1\linewidth,
        vertical sep = 0.06\linewidth,
    },
    title style={yshift=-1ex,},
    width=0.2\linewidth,
    height=0.2\linewidth,
    % labels, ticks, and ticklabels
    xtick={0,1,2},
    xticklabels={$0$,$0.5$,$1$},
    xlabel style={yshift=0.5ex,},
    ytick={0,1,2,3},
    yticklabels={$1$,$0.9$,$0.8$,$0.7$},
    % ESSENTIAL params for correct rendering and avoiding unnecessary whitespace
    y dir=reverse,
    enlargelimits={abs=0.5},
    xmin=0, xmax=2,
    ymin=0, ymax=3,      
    ]

    %% -----------------
    %% Top row (medians)
    %% -----------------
    % T maze (median)
    \nextgroupplot[
        title=T-maze (median),
        ylabel=Friction $\friction$,
        point meta min=\rtMinTmaze,
        point meta max=\rtMaxTmaze,    
        colormap name=bwrTmaze,       
    ]
        \addplot [matrix plot*,point meta=explicit] file {csv/rt/med_rt_heatmap_tmaze.txt};

    % Triangle (median)
    \nextgroupplot[
        title=Triangle (median),
        point meta min=\rtMinTriangle,
        point meta max=\rtMaxTriangle,
        colormap name=bwrTriangle,
    ]
        \addplot [matrix plot*,point meta=explicit] file {csv/rt/med_rt_heatmap_triangle.txt};

    % Biased rat (median)
    \nextgroupplot[
        title=Biased rat (median),
        point meta min=\rtMinBiased,
        point meta max=\rtMaxBiased,
        colormap name=bwrBiased,
        % Biased rat heatmap params
        xmax=3,
        xtick={0,1,2,3},
        xticklabels={$0$, $1$, $3$, $5$},
        yticklabels={$1$, $0.8$, $0.6$, $0.5$}
    ]
        \addplot [matrix plot*,point meta=explicit] file {csv/rt/med_rt_heatmap_biased.txt};

    %% ---------------------
    %% Bottom row (averages)
    %% ---------------------
    % T-maze (average)
    \nextgroupplot[
        title=T-maze (mean),
        point meta min=\rtMinTmaze,
        point meta max=\rtMaxTmaze,
        colormap name=bwrTmaze,  
        % Bottom row plot + heatmap params
        ylabel=Friction $\friction$,
        xlabel=Adaptation strength $b_a$,
        colorbar sampled, % ESSENTIAL FOR GROUPPLOTS
        colorbar horizontal,
        colorbar style={
            xtick={\rtMinTmaze, 0, \rtMaxTmazeHalf, \rtMaxTmaze},
            xticklabels={$\rtMinTmaze$, $0$, $+\rtMaxTmazeHalf$, $+\rtMaxTmaze$},
            xticklabel style={font=\small}
        }        
    ]
        \addplot [matrix plot*,point meta=explicit] file {csv/rt/avg_rt_heatmap_tmaze.txt};

    % Triangle (average)
    \nextgroupplot[
        title=Triangle (mean),
        point meta min=\rtMinTriangle,
        point meta max=\rtMaxTriangle,    
        colormap name=bwrTriangle,    
        % Bottom row plot + heatmap params
        xlabel=Adaptation strength $b_a$,
        colorbar sampled, % ESSENTIAL FOR GROUPPLOTS
        colorbar horizontal,
        colorbar style={
            xtick={\rtMinTriangle, 0, \rtMaxTriangleHalf, \rtMaxTriangle},
            xticklabels={$\rtMinTriangle$, $0$, $+\rtMaxTriangleHalf$, $+\rtMaxTriangle$},
            xticklabel style={font=\small}
        }           
    ]
        \addplot [matrix plot*,point meta=explicit] file {csv/rt/avg_rt_heatmap_triangle.txt};

    % Biased rat (average)
    \nextgroupplot[
        title=Biased rat (mean),
        point meta min=\rtMinBiased,
        point meta max=\rtMaxBiased,
        colormap name=bwrBiased,
        % Biased rat heatmap params
        xmax=3,
        xtick={0,1,2,3},
        xticklabels={$0$, $1$, $3$, $5$},
        yticklabels={$1$, $0.8$, $0.6$, $0.5$},
        % Bottom row plot + heatmap params
        xlabel=Adaptation strength $b_a$,
        colorbar sampled, % ESSENTIAL FOR GROUPPLOTS
        colorbar horizontal,
        colorbar style={
            xtick={\rtMinBiased, 0, \rtMaxBiasedHalf, \rtMaxBiased},
            xticklabels={$\rtMinBiased$, $0$, $+\rtMaxBiasedHalf$, $+\rtMaxBiased$},
            xticklabel style={font=\small}
        }        
    ]
        \addplot [matrix plot*,point meta=explicit] file {csv/rt/avg_rt_heatmap_biased.txt};

\end{groupplot}

\node[inner sep=0] at (-1.7,-4.5) {\column{Change (\%) from\\awake statistics:}};

\node[inner sep=0] at (4.2, 2.7)  {Replay reach time $(\downarrow)$ summary statistics};

\end{tikzpicture}
\caption{
This is another version of \Cref{fig:reachtime_heatmaps}, but now with mean reach times also shown.
}
\label{fig:reachtime_heatmaps_all}
\end{figure}

%% ------------------
%% Mean Displacements
%% ------------------

%% --------------------
%% Variances Along Time
%% --------------------

\input{tikzfigs/exploration_triangle}

\section{Results With Tanh Activations}
\label{appendix:tanh}

Throughout all other parts of the text, we have used piecewise-linear (linear or leaky/non-leaky ReLU) activation functions, as defended in \Cref{appendix:hyperparameters}.
For completeness, here we reproduce some key findings of our work with all hyperparameters kept the same, except that RNNs now use tanh activation functions instead of piecewise-linear ones.
Replay from tanh RNNs visited far fewer regions than replay from their piecewise-linear counterparts, so a reproduction of \Cref{fig:region_count_avg} (exploration results) is omitted.
Our primary finding is that \emph{underdampening induces temporal compression, and adaptation induces temporal dilation, in replay from tanh RNNs} (\Cref{fig:tanh_reachtime_heatmaps_all}).

%% ---------------------
%% Wasserstein distances
%% ---------------------
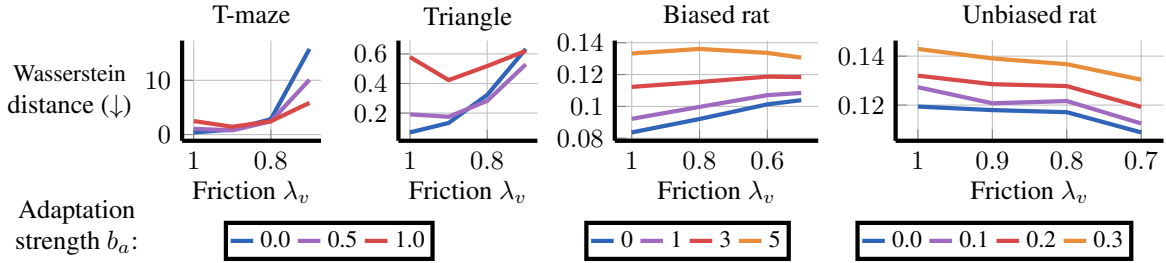
\begin{figure}[h]
\centering
\begin{tikzpicture}

\pgfplotsset{/pgfplots/group/every plot/.append style = {
    ultra thick,%, no markers
    legend style={
        legend columns=2,
        anchor=south west}
}};

\begin{groupplot}[
    group style = {
        group size = 4 by 1,
        horizontal sep = 0.06\linewidth},
    %width = 0.27\linewidth,
    width = 0.2\linewidth,
    height = 0.17\linewidth,
    title style={yshift=-1.3mm,},
    axis x line*=bottom,
    axis y line*=left,    
    x dir=reverse,   
    xlabel=Friction $\friction$,
    xlabel shift={-1mm},
    ylabel style={rotate=-90},
    ylabel shift={-3mm},
    yticklabel style={/pgf/number format/.cd,
    fixed},
    grid,
    no marks,
    table/col sep=comma,
    cycle list name=mygradient
    ]

    \nextgroupplot[title=T-maze,
        ylabel=\column{\small Wasserstein\\ distance ($\downarrow$)},
        legend style={
            legend columns=3,
            at={(0.3,-1.2)},
            inner sep=1pt,
            font=\small}]
            %font=\footnotesize}]
    \foreach \bval in {0.0,0.5,1.0}{
        \addplot table[x=lambda_v, y={b_a=\bval}] {csv/wd/tanh_wd_tmaze.csv}; 
        \addlegendentryexpanded{ \bval};
    }

    \nextgroupplot[
        title=Triangle,
        title style={yshift=-1mm,}]
    \foreach \bval in {0.0,0.5,1.0}{
        \addplot table[x=lambda_v, y={b_a=\bval}] {csv/wd/tanh_wd_triangle.csv}; 
    }

    \nextgroupplot[
        title={Biased rat},
        width=0.25\linewidth,
        legend style={
            legend columns=4,
            at={(-0.15,-1.2)},
            row sep=-3pt,
            inner sep=1pt,
            font=\small}]
            %font=\footnotesize}]
    \foreach \bval in {0,1,3,5}{
        \addplot table[x=lambda_v, y expr=\thisrow{b_a=\bval}] {csv/wd/tanh_wds_biased.csv}; 
        \addlegendentryexpanded{ \bval};
    }

    \nextgroupplot[
        title={Unbiased rat},
        width=0.3\linewidth,
        xtick={1, 0.9, 0.8, 0.7},
        legend style={
            legend columns=4,
            at={(-0.15,-1.2)},
            row sep=-3pt,
            inner sep=1pt,
            font=\small}]
            %font=\footnotesize}]
    \foreach \bval in {0.0,0.1,0.2,0.3}{
        \addplot table[x=lambda_v, y expr=\thisrow{b_a=\bval}] {csv/wd/tanh_wds_unbiased.csv};
        \addlegendentryexpanded{ \bval};
    }

\end{groupplot} 

\node[inner sep=0] at (-1.4,-1.15) {\column{Adaptation\\strength $b_a$:}};

\end{tikzpicture}
\caption{
\textbf{Mild underdampening can improve replay fidelity in the presence of adaptation in tanh RNNs.}
This is another version of \Cref{fig:wd}, for T-maze and triangle tasks, but with tanh RNNs.
}
\label{fig:tanh_wd}
\end{figure}

\newcommand{\tanhRtMinTmaze}{-30}
\newcommand{\tanhRtMaxTmazeHalf}{20}
\newcommand{\tanhRtMaxTmaze}{40}
\newcommand{\tanhRtMinTriangle}{-30}
\newcommand{\tanhRtMaxTriangleHalf}{45}
\newcommand{\tanhRtMaxTriangle}{90}
\newcommand{\tanhRtMinBiased}{-70}
\newcommand{\tanhRtMaxBiased}{0}
\pgfplotsset{
    colormap={tanhBwrTmaze}{
        color(\tanhRtMinTmaze)=(blue)
        color(0)=(white)
        color(\tanhRtMaxTmaze)=(red)},
    colormap={tanhBwrTriangle}{
        color(\tanhRtMinTriangle)=(blue)
        color(0)=(white)
        color(\tanhRtMaxTriangle)=(red)},
    colormap={tanhBwrBiased}{
        color(\tanhRtMinBiased)=(blue)
        color(\tanhRtMaxBiased)=(white)}
%        color(\tanhRtMaxBiasedHalf)=(white)
%        color(\tanhRtMaxBiased)=(red)}
}

%% -----------------------
%% all reach time heatmaps
%% -----------------------
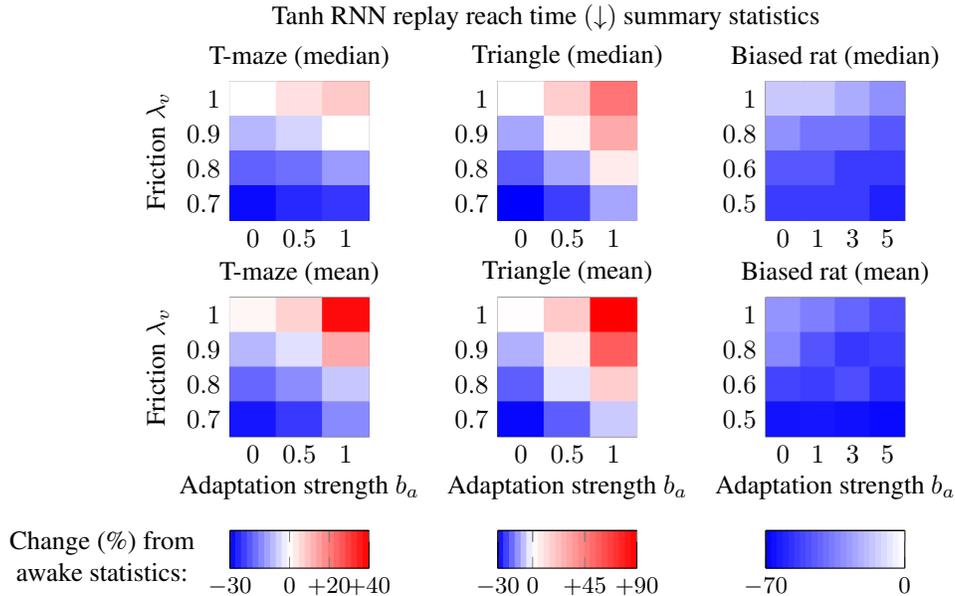
\begin{figure}[h]
\centering
\begin{tikzpicture}

\begin{groupplot}[
    group style = {
        group size = 3 by 2,
        horizontal sep = 0.1\linewidth,
        vertical sep = 0.06\linewidth,
    },
    title style={yshift=-1ex,},
    width=0.2\linewidth,
    height=0.2\linewidth,
    % labels, ticks, and ticklabels
    xtick={0,1,2},
    xticklabels={$0$,$0.5$,$1$},
    xlabel style={yshift=0.5ex,},
    ytick={0,1,2,3},
    yticklabels={$1$,$0.9$,$0.8$,$0.7$},
    % ESSENTIAL params for correct rendering and avoiding unnecessary whitespace
    y dir=reverse,
    enlargelimits={abs=0.5},
    xmin=0, xmax=2,
    ymin=0, ymax=3,      
    ]

    %% -----------------
    %% Top row (medians)
    %% -----------------
    % T maze (median)
    \nextgroupplot[
        title=T-maze (median),
        ylabel=Friction $\friction$,
        point meta min=\tanhRtMinTmaze,
        point meta max=\tanhRtMaxTmaze,    
        colormap name=tanhBwrTmaze,       
    ]
        \addplot [matrix plot*,point meta=explicit] file {csv/rt/tanh_med_rt_heatmap_tmaze.txt};

    % Triangle (median)
    \nextgroupplot[
        title=Triangle (median),
        point meta min=\tanhRtMinTriangle,
        point meta max=\tanhRtMaxTriangle,
        colormap name=tanhBwrTriangle,
    ]
        \addplot [matrix plot*,point meta=explicit] file {csv/rt/tanh_med_rt_heatmap_triangle.txt};

    % Biased rat (median)
    \nextgroupplot[
        title=Biased rat (median),
        point meta min=\tanhRtMinBiased,
        point meta max=\tanhRtMaxBiased,
        colormap name=tanhBwrBiased,
        % Biased rat heatmap params
        xmax=3,
        xtick={0,1,2,3},
        xticklabels={$0$, $1$, $3$, $5$},
        yticklabels={$1$, $0.8$, $0.6$, $0.5$}
    ]
        \addplot [matrix plot*,point meta=explicit] file {csv/rt/tanh_med_rt_heatmap_biased.txt};

    %% ---------------------
    %% Bottom row (averages)
    %% ---------------------
    % T-maze (average)
    \nextgroupplot[
        title=T-maze (mean),
        point meta min=\tanhRtMinTmaze,
        point meta max=\tanhRtMaxTmaze,
        colormap name=tanhBwrTmaze,  
        % Bottom row plot + heatmap params
        ylabel=Friction $\friction$,
        xlabel=Adaptation strength $b_a$,
        colorbar sampled, % ESSENTIAL FOR GROUPPLOTS
        colorbar horizontal,
        colorbar style={
            xtick={\tanhRtMinTmaze, 0, \tanhRtMaxTmazeHalf, \tanhRtMaxTmaze},
            xticklabels={$\tanhRtMinTmaze$, $0$, $+\tanhRtMaxTmazeHalf$, $+\tanhRtMaxTmaze$},
            xticklabel style={font=\small}
        }        
    ]
        \addplot [matrix plot*,point meta=explicit] file {csv/rt/tanh_avg_rt_heatmap_tmaze.txt};

    % Triangle (average)
    \nextgroupplot[
        title=Triangle (mean),
        point meta min=\tanhRtMinTriangle,
        point meta max=\tanhRtMaxTriangle,    
        colormap name=tanhBwrTriangle,    
        % Bottom row plot + heatmap params
        xlabel=Adaptation strength $b_a$,
        colorbar sampled, % ESSENTIAL FOR GROUPPLOTS
        colorbar horizontal,
        colorbar style={
            xtick={\tanhRtMinTriangle, 0, \tanhRtMaxTriangleHalf, \tanhRtMaxTriangle},
            xticklabels={$\tanhRtMinTriangle$, $0$, $+\tanhRtMaxTriangleHalf$, $+\tanhRtMaxTriangle$},
            xticklabel style={font=\small}
        }           
    ]
        \addplot [matrix plot*,point meta=explicit] file {csv/rt/tanh_avg_rt_heatmap_triangle.txt};

    % Biased rat (average)
    \nextgroupplot[
        title=Biased rat (mean),
        point meta min=\tanhRtMinBiased,
        point meta max=\tanhRtMaxBiased,
        colormap name=tanhBwrBiased,
        % Biased rat heatmap params
        xmax=3,
        xtick={0,1,2,3},
        xticklabels={$0$, $1$, $3$, $5$},
        yticklabels={$1$, $0.8$, $0.6$, $0.5$},
        % Bottom row plot + heatmap params
        xlabel=Adaptation strength $b_a$,
        colorbar sampled, % ESSENTIAL FOR GROUPPLOTS
        colorbar horizontal,
        colorbar style={
            xtick={\tanhRtMinBiased, \tanhRtMaxBiased},
            xticklabels={$\tanhRtMinBiased$, $\tanhRtMaxBiased$},
            xticklabel style={font=\small}
        }        
    ]
        \addplot [matrix plot*,point meta=explicit] file {csv/rt/tanh_avg_rt_heatmap_biased.txt};

\end{groupplot}

\node[inner sep=0] at (-1.7,-4.5) {\column{Change (\%) from\\awake statistics:}};

\node[inner sep=0] at (4.2, 2.7) {Tanh RNN replay reach time $(\downarrow)$ summary statistics};

\end{tikzpicture}
\caption{
\textbf{Underdampening increases speed in tanh RNN replay.}
This is another version of \Cref{fig:reachtime_heatmaps_all} for T-maze and triangle tasks, but now with tanh RNNs.
}
\label{fig:tanh_reachtime_heatmaps_all}
\end{figure}

%% -------------------
%% Exploration metrics
%% -------------------
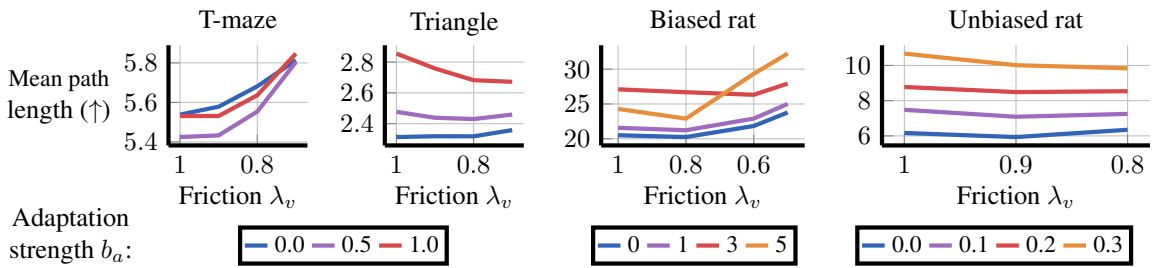
\begin{figure}[h!]
\centering
\begin{tikzpicture}

\pgfplotsset{/pgfplots/group/every plot/.append style = {
    ultra thick,%, no markers
    legend style={
        legend columns=2,
        anchor=south west}
}};

\begin{groupplot}[
    group style = {
        group size = 4 by 1,
        horizontal sep = 0.06\linewidth,
        vertical sep = 0.05\linewidth},
    %width = 0.27\linewith,
    width = 0.2\linewidth,
    height = 0.17\linewidth,
    title style={yshift=-1.3mm,},
    x dir=reverse,   
    axis x line*=bottom,
    axis y line*=left,
    xlabel={Friction $\friction$},
    xlabel shift={-1mm},    
    ylabel style={rotate=-90},
    ylabel shift={-3mm},
    grid,
    no marks,
    table/col sep=comma,
    cycle list name=mygradient
    ]

    % T-maze
    \nextgroupplot[
        title=T-maze,
        ylabel=\column{\small Mean path\\length ($\uparrow$)},
        legend style={
            legend columns=3,
            at={(0.5,-1.2)},
            inner sep=1pt,
            font=\small}]
    \foreach \bval in {0.0,0.5,1.0}{
        \addplot table[x=lambda_v, y={b_a=\bval}] {csv/exploration/tanh_exploration_tmaze_distances_avg.csv}; 
        \addlegendentryexpanded{ \bval};
    }

    % Triangle
    \nextgroupplot[
        title=Triangle,
        title style={yshift=-1mm,},]
    \foreach \bval in {0.0,0.5,1.0}{
        \addplot table[x=lambda_v, y={b_a=\bval}] {csv/exploration/tanh_exploration_triangle_distances_avg.csv};  
    }

    % Biased rat
    \nextgroupplot[
        title=Biased rat,
        width=0.25\linewidth,
        legend style={
            at={(-0.05,-1.2)},
            legend columns=4,
            row sep=-3pt,
            inner sep=1pt,
            font=\small}]
    \foreach \bval in {0,1,3,5}{
        \addplot table[x=lambda_v, y={b_a=\bval}] {csv/exploration/tanh_exploration_distance_biased.csv};  
        \addlegendentryexpanded{ \bval};
    }    

    % Unbiased rat
    \nextgroupplot[
        title=Unbiased rat,
        width=0.3\linewidth,
        legend style={
            at={(-0.1,-1.2)},
            legend columns=4,
            row sep=-3pt,
            inner sep=1pt,
            font=\small}]
    \foreach \bval in {0.0,0.1,0.2,0.3}{
        \addplot table[x=lambda_v, y={b_a=\bval}] {csv/exploration/tanh_exploration_distance_unbiased_edited.csv};  
        \addlegendentryexpanded{ \bval};
    }

\end{groupplot} 

\node[inner sep=0] at (-1.3,-1.2) {\column{Adaptation\\strength $b_a$:}};

\end{tikzpicture}
\caption{
\textbf{In tanh RNNs, underdampening may increase or decrease exploration via path length.}
This is an alternate version of \Cref{fig:exploration_metrics} but with tanh RNNs.
Underdampening has mixed effects on tanh RNN replay path length because of tanh saturation.
}
\label{fig:tanh_exploration_metrics}
\end{figure}

\end{document}